\def\paperlanguage{} 

\documentclass[letterpaper, 10 pt, conference]{ieeeconf}  

\usepackage{bm}
\usepackage{cite}
\usepackage{flushend}
\include{preamble}

\IEEEoverridecommandlockouts                              

\title{\LARGE \textbf
  {
    \switchlanguage%
    {%
      An RGB-D Camera-Based
      Multi-Small Flying Anchors Control \\
      for Wire-Driven Robots Connecting to the Environment
    }%
    {%
      An RGB-D Camera-Based Environment Recognition \\
      and Multi-Small Flying Anchors Control \\
      for Wire-Driven Robots Connecting to the Environment
    }%
  }
}

\author{Shintaro Inoue$^{1}$, Kento Kawaharazuka$^{1,2}$, Keita Yoneda$^{1}$, Sota Yuzaki$^{1}$, \\Yuta Sahara$^{1}$, Temma Suzuki$^{1}$, and Kei Okada$^{1}$
  \thanks{$^{1}$ The authors are with the Department of Mechano-Informatics, Graduate School of Information Science and Technology, The University of Tokyo, 7-3-1 Hongo, Bunkyo-ku, Tokyo, 113-8656, Japan.
    {\texttt\small [s-inoue, kawaharazuka, yoneda, yuzaki, sahara, t-suzuki, k-okada]@jsk.imi.i.u-tokyo.ac.jp}
  }
  \thanks{$^{2}$ The author is with the AI Center, Graduate School of Information Science and Technology, The University of Tokyo, Japan.}
}
\begin{document}

\maketitle
\thispagestyle{empty}
\pagestyle{empty}

\begin{abstract}
  \switchlanguage%
  {%
    In order to expand the operational range and payload capacity of robots, 
    wire-driven robots that leverage the external environment have been proposed. 
    It can exert forces and operate in spaces far beyond those dictated by its own structural limits. 
    However, for practical use, 
    robots must autonomously attach multiple wires to the environment based on environmental recognition—an operation 
    so difficult that many wire-driven robots remain restricted to specialized, pre-designed environments.
    Here, in this study, we propose a robot that autonomously connects multiple wires to the environment 
    by employing a multi-small flying anchor system, 
    as well as an RGB-D camera-based control and environmental recognition method.
    Each flying anchor is a drone with an anchoring mechanism at the wire tip, 
    allowing the robot to attach wires by flying into position. 
    Using the robot's RGB-D camera to identify suitable attachment points and a flying anchor position, 
    the system can connect wires in environments that are not specially prepared, 
    and can also attach multiple wires simultaneously.
    Through this approach, a wire-driven robot can autonomously attach its wires to the environment, 
    thereby realizing the benefits of wire-driven operation at any location.
  }%
  {%
    ロボットの活動範囲やペイロードを増加させる手段として，ロボット外部の環境を利用したワイヤ駆動ロボットが提案されてきた．
    ロボットに搭載されたウインチが，
    ロボットと外部環境を接続したワイヤを巻き取ることで張力を発生させ， その張力でロボットを駆動する．
    ワイヤ破断強度までの強い力を，数メートルオーダーのワイヤ長さという広い範囲で利用できるため，
    ロボットの身体構造から想定されるよりも遥かに大きな力と空間が扱えるようになる．
    しかし、このようなロボットを利用するためには，
    ロボット外部の環境を認識した上で複数本のワイヤを接続するという工程が不可欠であり，
    これを自律実行する難易度の高さから，多くのワイヤ駆動ロボットは予めワイヤが張れるよう設計された限定的な環境内でしか，
    その能力を発揮できていない．
    そこで、本研究では、マルチ小型飛行アンカーと，RGBDカメラを用いたその制御と環境認識手法により，
    自律的に環境に複数のワイヤを接続し駆動できるロボットを提案する．
    飛行アンカーは，アンカーを搭載したドローンで，各ワイヤの先端に取り付けらており，
    自身が飛行することで環境にワイヤを結びつける．
    また，飛行アンカーとワイヤを接続する環境をロボットに搭載されたRGBDカメラで認識しながら，
    ワイヤの結びつけを行うことで，予め定められた環境に留まらず，屋外環境でのワイヤ接続，複数本同時のワイヤ接続を実現した．
    これにより、ワイヤ駆動ロボットが活動するべき場所で自律的にワイヤを環境に接続し，ワイヤ駆動を活かした活躍を実現する．
  }%
\end{abstract}

\section{Introduction}\label{sec:introduction}
\switchlanguage%
{%

  As a means of expanding a robot's operational range, payload, and the variety of achievable tasks, 
  robots that connect themselves to external environments have been developed \cite{5979640,10619924,10296064,titan_xi_slope,8794265,9517518,10375200,5509299,merlet2010marionet,cubix}. 
  For example, some robots use magnets to attach themselves to ferromagnetic materials 
  and thereby achieve various locomotion methods \cite{5979640,10619924}, 
  while others attach themselves to flat surfaces by suction generated through rotors, 
  enabling a transformable flying robot to perform tasks \cite{10296064}. 
  Another example is a quadruped robot that uses wires to connect itself to the environment, 
  enabling stable walking on slopes while carrying out tasks \cite{titan_xi_slope}.

  Focusing specifically on robots that attach to the environment via wires, 
  these robots extend wires from themselves to the external environment 
  and generate tension using onboard winches that wind the wires.
  By combining the external environment with the mechanical properties of the wires, 
  these robots can harness powerful forces (up to the wire's breaking strength) over lengths on the order of several meters. 
  Consequently, 
  wire-driven robots that leverage their surrounding environment can handle forces and operate within spaces 
  far beyond what their own structures would typically allow.
  \begin{figure}[t]
    \begin{center}
      \includegraphics[width=1.0\columnwidth]{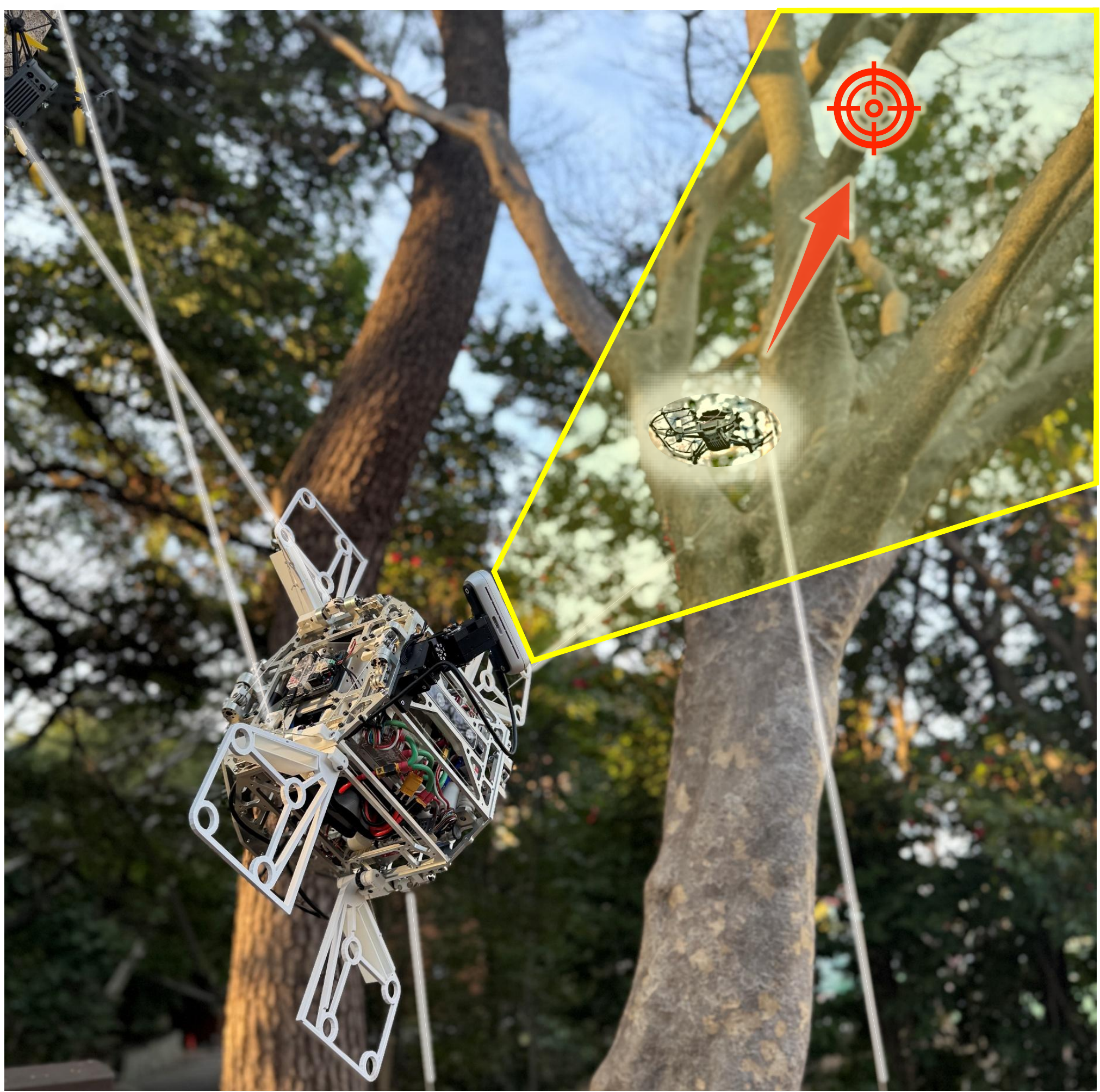}
      \vspace{-4.0ex}
      \caption{
        The wire-driven robot CubiX and a flying anchor that connect a wire to the environment 
        using RGB-D camera-based recognition.
      }
      \vspace{-4.0ex}
      \label{fig:fig1}
    \end{center}
  \end{figure}

  One such example is a system in which a drone and a vehicle are connected by a wire 
  to improve the vehicle's mobility \cite{8794265}. 
  Another study aiming to increase a drone's payload \cite{9517518} employed two child drones attached by wires 
  to a parent drone. 
  Furthermore, research on increasing the payload and output force of a wheeled humanoid robot \cite{10375200} 
  used a powered carabiner to attach a wire to pipes in the environment.
  In a more practical scenario, 
  the MARIONET-CRANE system \cite{5509299,merlet2010marionet}—equipped with wire-winding modules—relies on 
  human operators to manually connect wires to objects in the environment (e.g., debris), 
  enabling rescue operations.
  In all of these examples, by leveraging the environment through wire connections, 
  the robots enhance their mobility, payload capabilities, or output forces.

  However, operating this type of robot 
  demands the ability to recognize attachment points in the environment and connect multiple wires 
  to them—an autonomous process that remains challenging. 
  In prior work, limitations have persisted, such as constraints on the shapes of hooks or carabiners 
  that restrict attachable environments \cite{9517518,10375200}, 
  confining the number of attachable wires to just one \cite{8794265}, 
  or requiring human intervention to make the wire connections \cite{5509299,merlet2010marionet}.

  Therefore, in this study, we address these challenges by using the wire-driven robot CubiX \cite{cubix}, 
  which is equipped with an RGB-D camera, together with multiple small flying anchors (\figref{fig:fig1}). 
  CubiX houses wire-winding modules internally and is driven by winding multiple wires 
  that extend from within its body to the environment. 
  The flying anchors, which are drones outfitted with anchoring mechanisms, 
  are attached to the tip of these wires. 
  By using CubiX's RGB-D camera to recognize both the environment and the flying anchors, 
  we can control each anchor and attach the wire to the environment.

  As a result, our system can attach wires not only in predefined indoor environments but also in outdoor settings, 
  and can connect multiple wires at once. 
  Through this approach, wire-driven robots can autonomously attach their wires to the environment,
  thereby forming and exploiting wire-driven motion in a self-directed manner.
}%
{%
  \begin{figure}[t]
    \begin{center}
      \includegraphics[width=1.0\columnwidth]{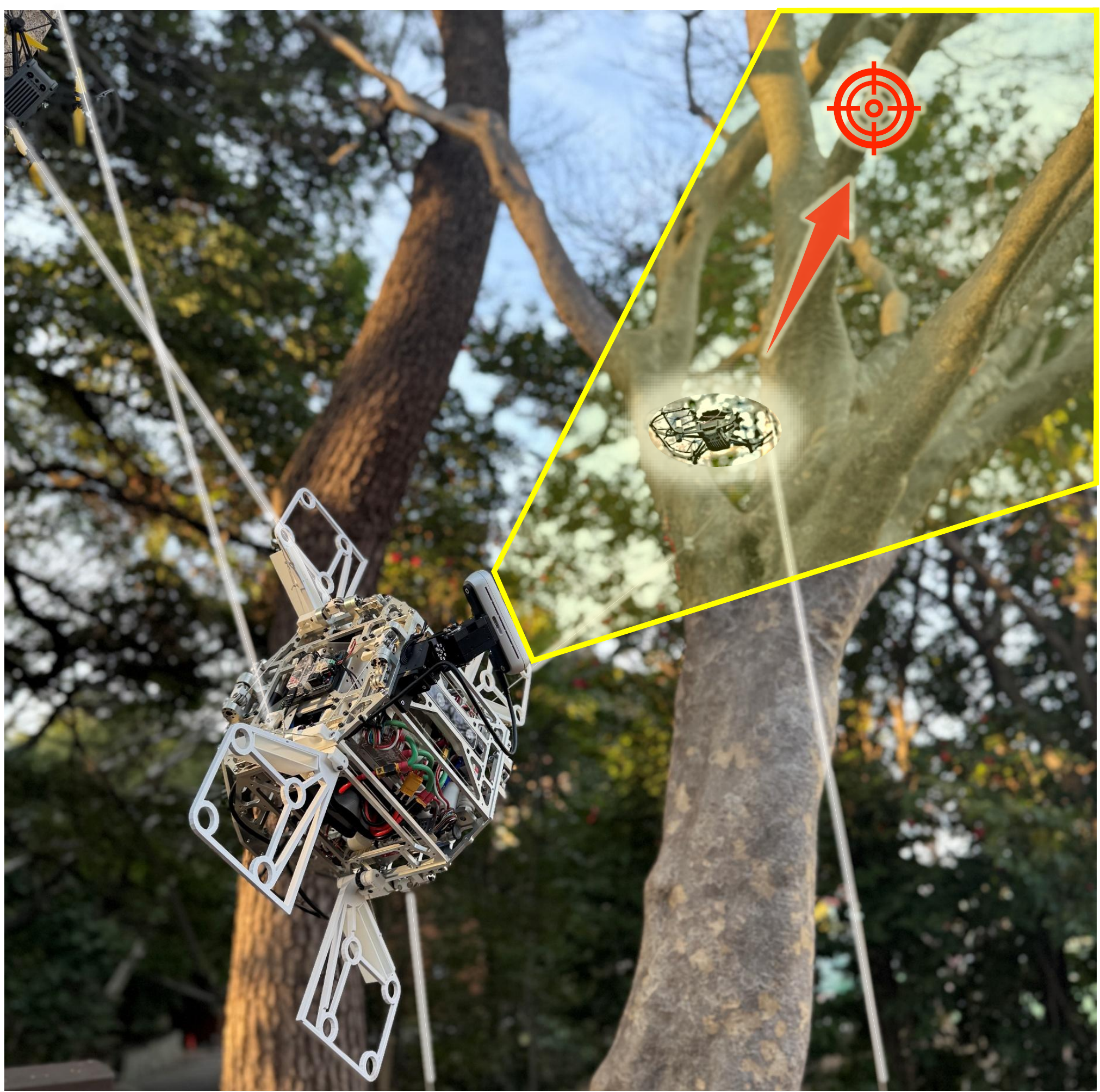}
      \vspace{-5.0ex}
      \caption{ワイヤ駆動ロボットCubiXとRGBDカメラによる観測を用いてワイヤを環境に接続する飛行アンカー．
      }
      \vspace{-7.0ex}
      \label{fig:fig1}
    \end{center}
  \end{figure}
  ロボットの活動範囲やペイロード，実現可能なタスクを拡大させる手段として，
  ロボット外部の環境にロボット自身を接続させて，環境を利用するロボットが開発されてきた\cite{5979640,10619924,10296064,titan_xi_slope,8794265,9517518,10375200,5509299,merlet2010marionet,cubix}．
  例えば，磁石を用いて磁性体に自身を接続し，多様なロコモーションを形成するロボット\cite{5979640,10619924}や，
  ロータによる吸引で平面環境に自身を接続することで，タスクを遂行する変形飛行ロボット\cite{10296064}，
  ワイヤを用いて環境に自身を接続し，斜面を安定して歩行しながら作業する四脚ロボット\cite{titan_xi_slope}などが開発されてきた．

  特に，ワイヤを介した環境接続を行うロボットに注目すると，
  ロボットは自身から出るワイヤをロボット外部の環境に接続し，
  ロボットに搭載されたウインチでそれらのワイヤを巻き取ることによって発生させた張力で駆動する．
  ロボット外部の環境とワイヤの特性を組み合わせることで，
  ワイヤの破断強度までの強い力を，数メートルオーダーのワイヤ長さという広い範囲で利用できる．
  これにより，環境を利用するワイヤ駆動ロボットは，
  その身体構造から想定されるよりも遥かに大きな力と空間を扱うことができる．

  そのようなロボットの例として，
  ワイヤで接続されたドローンと車両が連携することで車両の走破性を向上させる研究\cite{8794265}がある．
  この研究では，車両にワイヤで接続された1台のドローンが環境にワイヤを接続し，
  車両がそのワイヤを巻き取ることで，崖や階段などの車両だけでは移動できない場所での移動を可能にした．
  また，ドローンのペイロードを増加させる研究\cite{9517518}では，
  親ドローンにワイヤで接続された子ドローン2台が，フックで環境に引っ掛かり，
  親ドローンがそれらのワイヤを巻き取ることで，親ドローン本来のペイロードを増加させている．
  さらに，台車型ヒューマノイドロボットのペイロードや発揮力を増加させる研究\cite{10375200}では，
  駆動可能なロック機構付きのカラビナをパイプを用いて環境に引っ掛け，
  台車型ヒューマノイドロボットがそのワイヤを巻き取ることで，ペイロードや発揮力を増加させている．
  より実用的な場面を想定した研究として，MARIONET-CRANE\cite{5509299,merlet2010marionet}では，
  ワイヤ巻取りモジュールを搭載したクレーンから出るワイヤを人力で環境に存在する瓦礫などの対象物に接続することで，
  瓦礫撤去を含む救助活動を実現している．
  これらのロボットはいずれも，ロボット自身と環境とをワイヤで接続することで環境を利用することで，
  ロボットが強い力を広い範囲で扱えるようになり，自身の移動性能やペイロード，発揮力を向上させている．
  
  一方で，ワイヤを介して環境を利用するこのようなロボットを扱うためには，
  ロボットが環境からワイヤ接続箇所を認識し，複数のワイヤをそこらへ接続するという工程が必須となり，
  これを自律的に実現することは高い難易度を持つ．
  ロボット外部に存在する環境は，設計できない未知のものであり，
  様々な環境に対してワイヤを接続できるような仕組みであることが求められる．
  さらに，ワイヤの本数が複数本の場合でもシステムが稼働できる必要もある．
  先述した先行研究についても，フックやカラビナの形状で接続できる環境が制限されている\cite{9517518,10375200}，
  接続するワイヤの本数が1本に制限されている\cite{8794265}，
  そもそも人力でのみワイヤを接続している\cite{5509299,merlet2010marionet}
  といったように，課題が残されてきた．

  そこで，本研究では，
  \figref{fig:fig1}に示されるRGBDカメラを搭載した環境接続ワイヤ駆動ロボットCubiX\cite{cubix}と，マルチ小型飛行アンカーにより，
  自律的に複数のワイヤを環境に接続し，ロボットが駆動することを実現する．
  CubiXはワイヤを巻き取るモジュールをロボット体内に内蔵し,
  ロボット体内から出る複数のワイヤを環境に接続させ, それを巻き取ることで駆動する.
  飛行アンカーは，アンカーを搭載したドローンであり，CubiXの各ワイヤの先端に取り付けられている．
  CubiXに搭載されたRGBDカメラを用いて環境からワイヤ接続箇所を認識しながら，
  飛行アンカーもCubiXから観測しつつ制御する．
  その結果，予め定められた環境に限らず，屋外の自然環境へのワイヤ接続が可能となり，
  また，複数本のワイヤを同時に環境へ接続することも実現した．
  これにより，ワイヤ駆動ロボットが活動するべき場所で，自律的にワイヤを環境へ接続し，
  ワイヤ駆動を形成できる．



}%

\section{Hardware} \label{sec:hardware}
\switchlanguage%
{%
  \begin{figure}[t]
    \begin{center}
      \includegraphics[width=1.0\columnwidth]{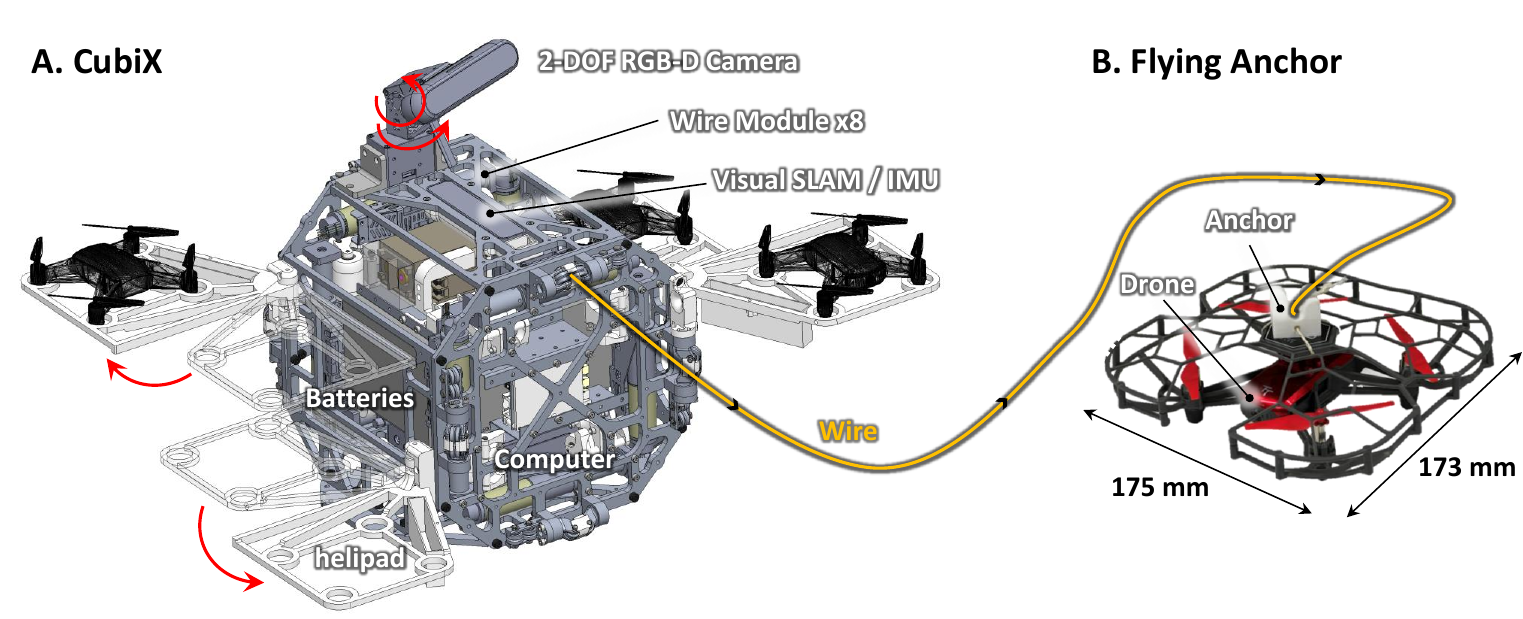}
      \vspace{-5.0ex}
      \caption{
        Overview of CubiX and the flying anchors. 
        CubiX is a wire-driven robot equipped with eight wire-winding modules and a two Degrees of Freedom (DOF) RGB-D camera.
        Each wire's endpoint has a flying anchor comprising a drone and anchoring mechanism. 
      }
      \vspace{-5.0ex}
      \label{fig:hardware}
    \end{center}
  \end{figure}

  \figref{fig:hardware} shows an overview of the wire-driven robot CubiX and the flying anchors. 
  CubiX is a robot equipped with actuators to wind wires internally;
  it can connect up to eight wires to the environment and then wind them to move. 
  Each wire module provides a maximum continuous tension of 180 N
  and a total winding length of 5.3 m.
  CubiX has the computer, batteries, and a V-SLAM camera (RealSense T265). 
  On top, it is equipped with an RGB-D camera (RealSense D455) that can pivot in yaw and pitch. 

  Two of the robot's side faces each have space for two helipads, on which the flying anchors are docked when not in flight.
  Each helipad carries a flying anchor, which is a drone fitted with an anchoring mechanism on top. 
  To allow for maneuvering through narrow spaces when attaching wires to the environment, 
  each drone is a lightweight, compact Tello EDU \cite{TelloEDU}. 
  The attached anchor is a 3D-printed component that withstands compressive forces during wire tension; 
  it is made of PLA, weighs only 5 g, yet is verified to support loads of up to 340 N. 
}%
{%
  \begin{figure}[t]
    \begin{center}
      \includegraphics[width=1.0\columnwidth]{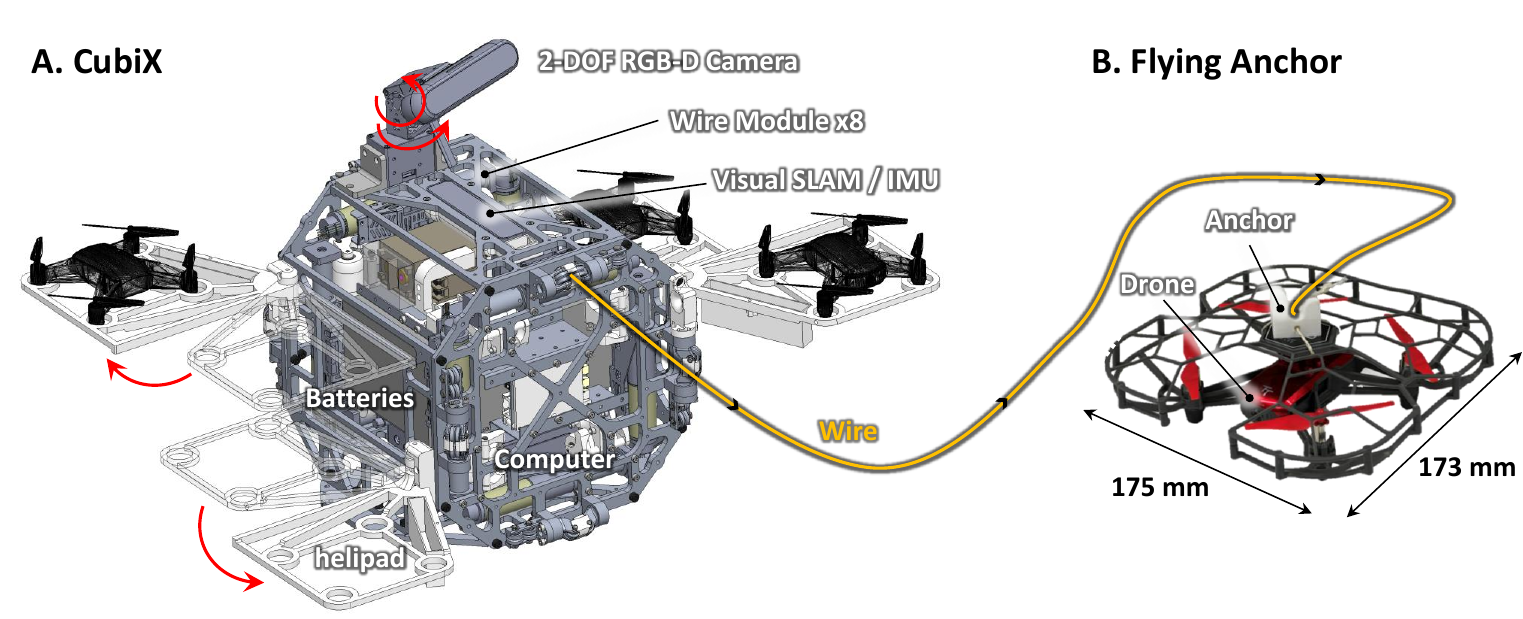}
      \vspace{-5.0ex}
      \caption{CubiXと飛行アンカーの全体像．CubiXはワイヤを巻き取るモジュールが8台搭載されたワイヤ駆動ロボットで，
      そのワイヤの先端には，ドローンとアンカーを組み合わせた飛行アンカーが取り付けられている．
      また，CubiXには環境や飛行アンカーを観測する2自由度RGBDカメラが搭載されている．
      }
      \vspace{-5.0ex}
      \label{fig:hardware}
    \end{center}
  \end{figure}

  本研究で用いるワイヤ駆動ロボットCubiXと飛行アンカーの全体像を\figref{fig:hardware}に示す．
  CubiXは，ワイヤを巻き取るアクチュエータをロボット体内に内蔵し，
  ロボット体内から出る最大8本のワイヤを環境に接続させ，それを巻き取ることで駆動するロボットである．
  キューブ形状の身体構造の辺に沿ってワイヤを巻き取るワイヤモジュールが合計で8個，組み付けられている．
  ワイヤモジュールは最大連続張力 18 kgf，ワイヤ巻取り速度 242 mm/s, ワイヤ巻取り長さ 5.3 m という性能を持つ．
  また，コンピュータ，バッテリ，V-SLAMカメラ（Intel Realsense T265）といった稼働に必要なデバイスをその体内に搭載しており，
  CubiXは自律して活動することができる．
  上面には，yaw方向とpitch方向に可動なRGBDカメラ（Intel Realsense D455）が搭載されており，
  ロボット外部環境におけるワイヤ取り付け箇所の認識や，飛行アンカーの位置観測を担う．
  側面2面には，飛行アンカーが搭載されるスペースであるhelipadを各2台搭載している．
  このhelipadは，通常時は折りたたまれた状態であるが，飛行アンカーが離陸する際には左右に展開され，
  4台の飛行アンカーが干渉することなく離陸できる．
  本研究では，この4台のhelipadを駆動するために，8本あるワイヤの内4本を使用している．
  そのhelipadの上に，CubiXとワイヤで接続された飛行アンカーが搭載されている．
  飛行アンカーは，ドローンの上部にアンカーを取り付けたものである．
  飛行アンカーは，環境にワイヤを接続する際に狭隘空間の通過を求められることがあるため，
  そのドローンには軽量かつコンパクトなDJI Tello EDU\cite{TelloEDU}を使用した．
  アンカーは環境接続時のワイヤ張力を圧縮方向に受ける3Dプリンタパーツであり，
  PLA製で5 gと軽量でありながら35 kgの荷重に耐えることが確認されている．
  本研究では，CubiXと，4台の飛行アンカーを用いる．
}%

\section{Software} \label{sec:software}
\subsection{System Configuration}
\switchlanguage%
{%
  \begin{figure}[t]
    \begin{center}
      \includegraphics[width=1.0\columnwidth]{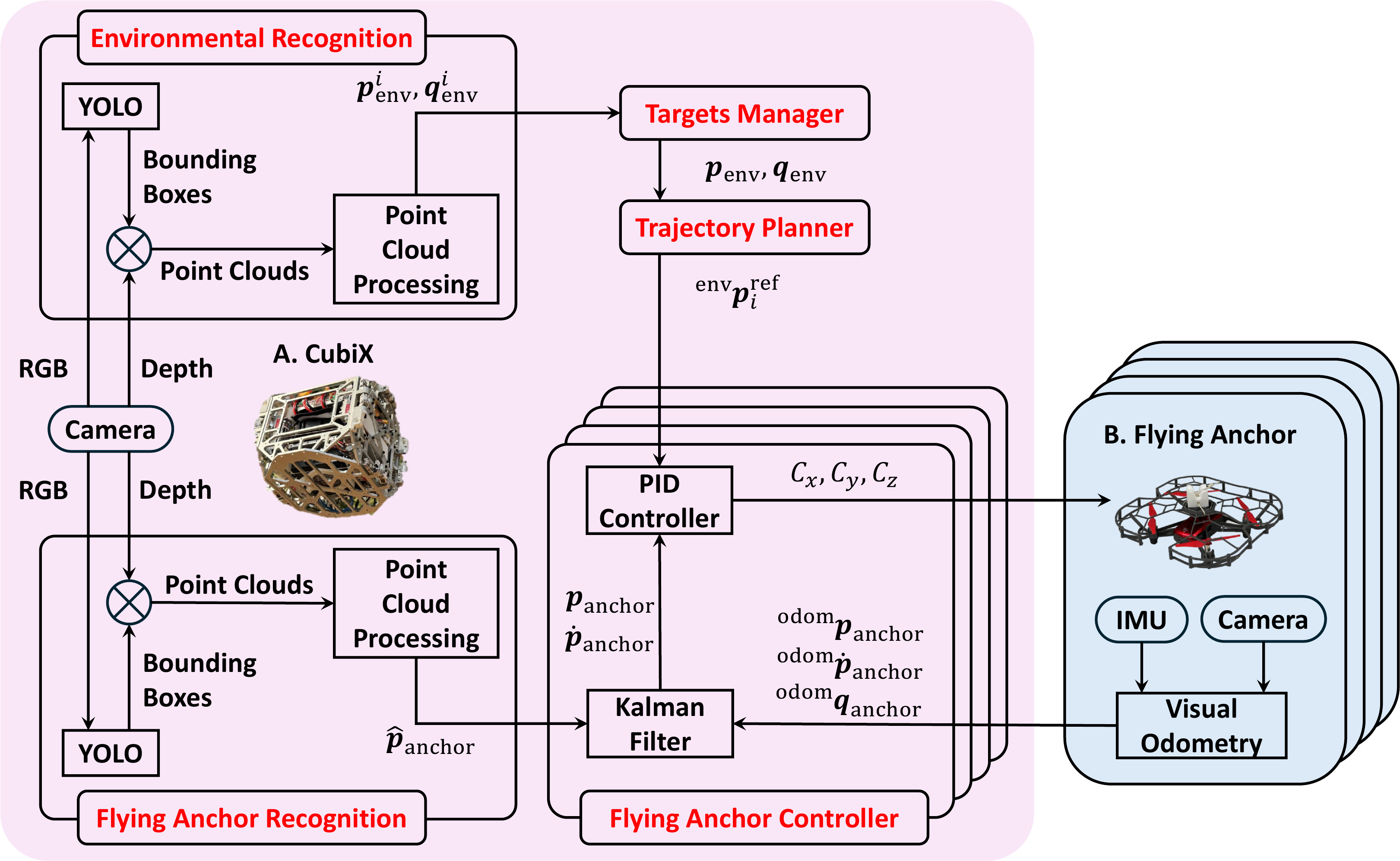}
      \vspace{-5.0ex}
      \caption{
        System configuration diagram. 
        CubiX's RGB-D camera recognizes both the environment and the flying anchors, 
        and controls multiple flying anchors simultaneously.
      }
      \vspace{-5.0ex}
      \label{fig:software}
    \end{center}
  \end{figure}

  \figref{fig:software} shows the overall configuration of the system, 
  which consists of Environmental Recognition, Targets Manager, Trajectory Planner, Flying Anchor Recognition, 
  and Flying Anchor Controller. 
  In Environmental Recognition, the RGB-D camera is used for recognition of potential wire attachment targets.
  Targets Manager then manages multiple recognized target positions for wire attachment. 
  Next, Trajectory Planner 
  generates a trajectory for the flying anchors to attach the wire. 
  In Flying Anchor Recognition, the RGB-D camera is used for recognition of the flying anchors' positions.
  Finally, Flying Anchor Controller controls each flying anchor to follow the generated trajectory, 
  running in parallel for every flying anchor.
  These processes run on CubiX, as shown in \figref{fig:software}A. 
  On each flying anchor (\figref{fig:software}B), visual odometry is used internally to estimate the anchor's position 
  in the \textit{odom} coordinate frame. 
  CubiX broadcasts a 2.4 GHz Wi-Fi network as an access point, 
  and all flying anchors communicate with CubiX through this network.

  A possible alternative for achieving the same objective would be a system with no cameras on CubiX, 
  relying instead on the flying anchors' cameras for both environmental recognition and self-localization. 
  However,
  each flying anchor must be compact and cannot accommodate extensive computational resources, 
  making onboard image processing impractical. 
  Although one might consider transmitting camera images from the flying anchors to CubiX for processing, 
  such an approach could introduce significant lag or delay due to Wi-Fi congestion, 
  especially when multiple anchors are deployed.

}%
{%
  \begin{figure}[t]
    \begin{center}
      \includegraphics[width=1.0\columnwidth]{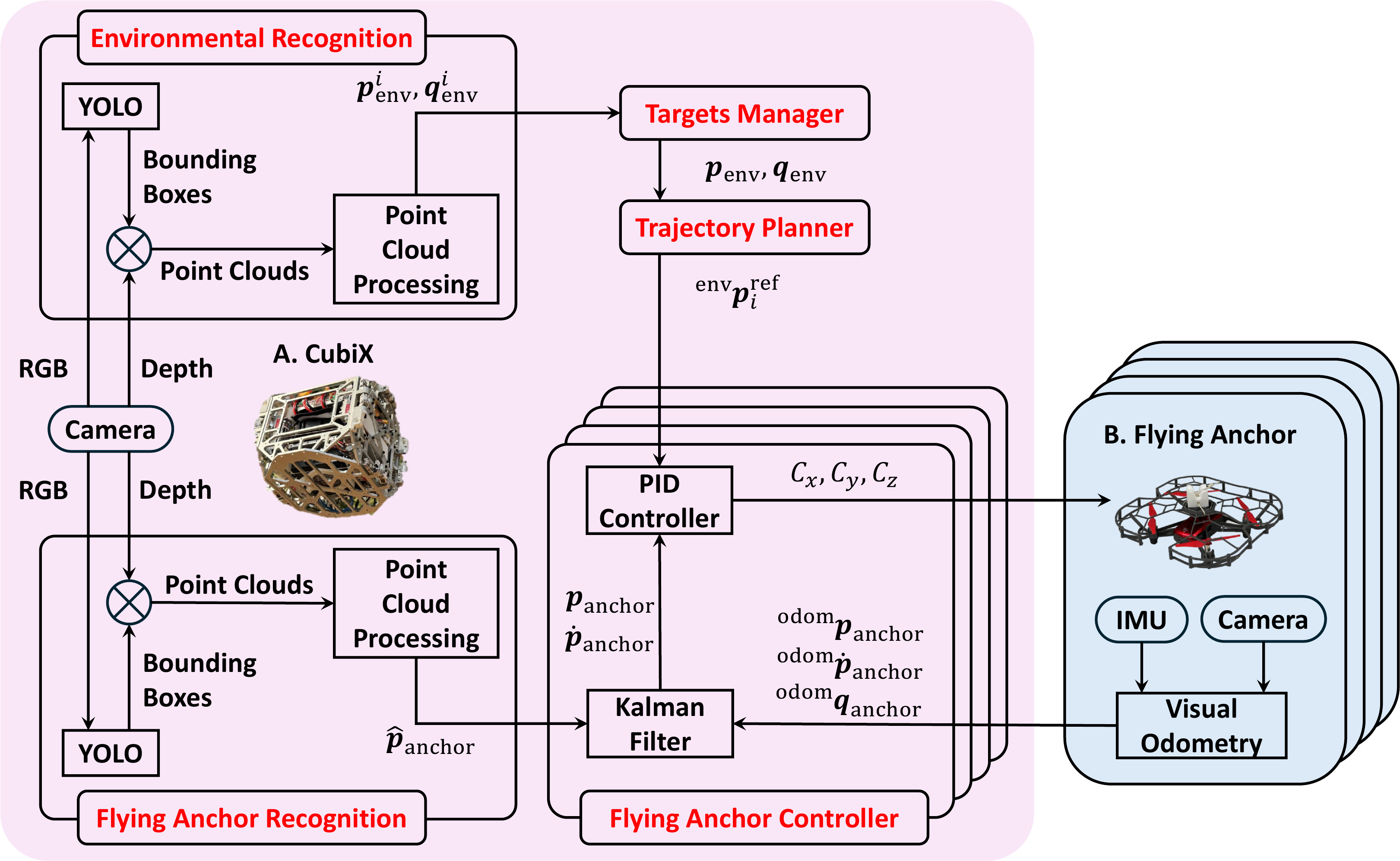}
      \vspace{-5.0ex}
      \caption{システム構成図．CubiXのRGBDカメラで環境と飛行アンカーを認識し，複数台の飛行アンカーを制御する．
      }
      \vspace{-5.0ex}
      \label{fig:software}
    \end{center}
  \end{figure}
  本研究のシステム全体の構成図を\figref{fig:software}に示す．
  システムは，
  environmental perception，targets manager，trajectory planner，flying anchor perception，flying anchor contollerからなる．
  environmental perceptionでは，RGBDカメラを用いて，CubiX外部の環境からワイヤ接続箇所の目標位置を認識する．
  targets managerでは，そこで認識された複数のワイヤ接続目標位置をカルマンフィルターを用いて管理する．
  trajectory plannerでは，指定された目標位置相対で，ワイヤを環境に結びつけるための飛行アンカーの目標軌跡を生成する．
  flying anchor perceptionでは，RGBDカメラを用いて，世界座標系における飛行アンカーの位置を観測する．
  flying anchor controllerは，飛行アンカーが生成された経路を追従するための制御を行うもので，
  飛行アンカーの台数分，パラレルに実行される．制御量は飛行アンカーに張り付いた座標系における，並進方向の0から100の操作量である．
  上記の処理が\figref{fig:software}AのCubiX上で実行される．
  \figref{fig:software}Bの飛行アンカーでは，内部実行されるビジュアルオドメトリを用いて，
  odom座標系における自身の自己位置を推定する．
  なお，CubiXがアクセスポイントとして2.4GHz帯のWi-Fiを発信し，
  CubiXと全ての飛行アンカーはそのネットワークを介して通信を行う．

  本研究と同様の目的を満たすシステムとして，CubiXにカメラを搭載せず，
  飛行アンカーのカメラを用いて，環境認識や自身の位置観測を行う枠組みも考えられる．
  しかし，先述した通り飛行アンカー自体はコンパクトであることが求められるため，豊富な計算資源を積むことはできない．
  そのため，飛行アンカー自身が映像処理を行うことはできない．
  また，飛行アンカーのカメラ画像をCubiXに転送し，CubiXが画像処理を行うシステムも考えられるが，
  飛行アンカーが複数台になった場合について，画像を通信することはラグや遅延を招くため避けられたい．
  よって，本研究ではCubiXがカメラを持ち，環境と飛行アンカーの両方を認識するというシステムを採用している．
}%

\subsection{Environmental Recognition} \label{sec:env_recognition}
\switchlanguage%
{%

 
  \figref{fig:env_perception} shows an overview of Environmental Recognition. 
  First, YOLO\cite{7780460} (YOLO11) is used to obtain bounding boxes for potential wire attachment locations 
  from the RGB image captured by CubiX's camera (\figref{fig:env_perception}-1). 
  The YOLO inference model was pre-trained using manually annotated images of the frames employed in our experiments, 
  along with horizontally oriented tree branches, labeled via CVAT\cite{boris_sekachev_2020_4009388}.
  Using the obtained bounding boxes, we then mask the depth image captured at the same time. 
  During this masking step, we reduce the area inside each bounding box so that we eliminate information 
  unrelated to the target object.
  Next, a point cloud is generated using the masked depth image and the camera parameters 
  (\figref{fig:env_perception}-2).
  Subsequently, the position and orientation of the wire attachment point 
  are determined from the generated point cloud (\figref{fig:env_perception}-3). 
  Since the point cloud contains noise, 
  we first perform downsampling and then select the nearest cluster using Euclidean clustering. 
  We compute the centroid of that cluster and perform principal component analysis (PCA) \cite{jolliffe2002principal} 
  to find the direction in which the point cloud is dispersed. 
  We use PCL\cite{5980567}, a point cloud processing library, for the downsampling and clustering steps.
  We define a \textit{target} coordinate frame whose origin is at this centroid, 
  whose $y$-axis aligns with the main principal component, whose $z$-axis lies on the vertical plane.
  We denote the position and orientation of this \textit{target} coordinate frame in the world coordinate system 
  by $\bm{p}^{i}_\text{target}, \,\bm{q}^{i}_\text{target}$. 
  Because there can be as many \textit{target} coordinate frames as bounding boxes, 
  the index $i$ is used to distinguish them. 

  \begin{figure}[t]
    \begin{center}
      \includegraphics[width=1.0\columnwidth]{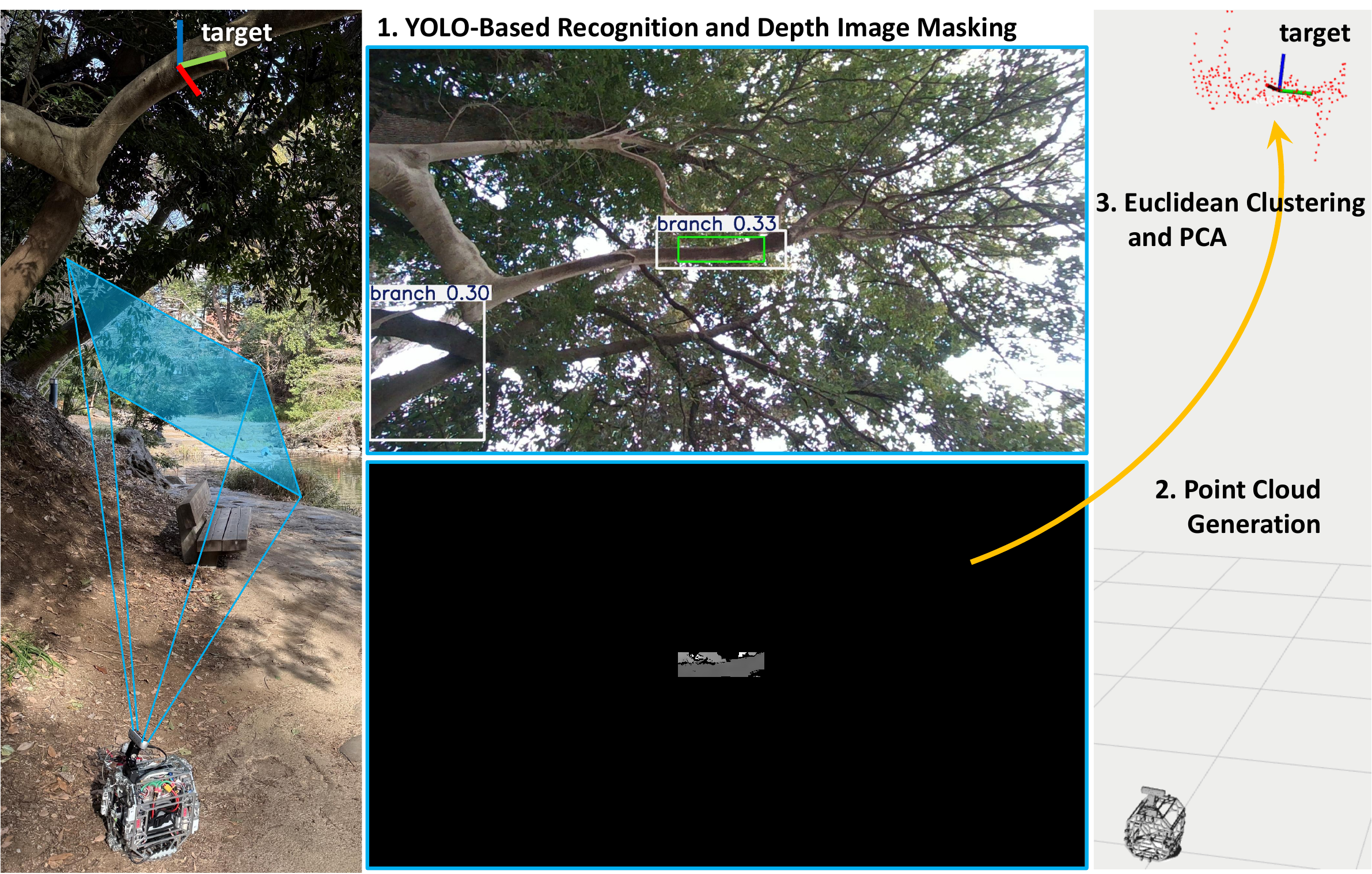}
      \vspace{-5.0ex}
      \caption{
        Overview of Environmental Recognition. 
        A point cloud is generated from the depth image, 
        masked at potential wire-attachment points, 
        and the position and orientation of the attachment target are determined.
      }
      \vspace{-4.0ex}
      \label{fig:env_perception}
    \end{center}
  \end{figure}
}%
{%
  \begin{figure}[t]
    \begin{center}
      \includegraphics[width=1.0\columnwidth]{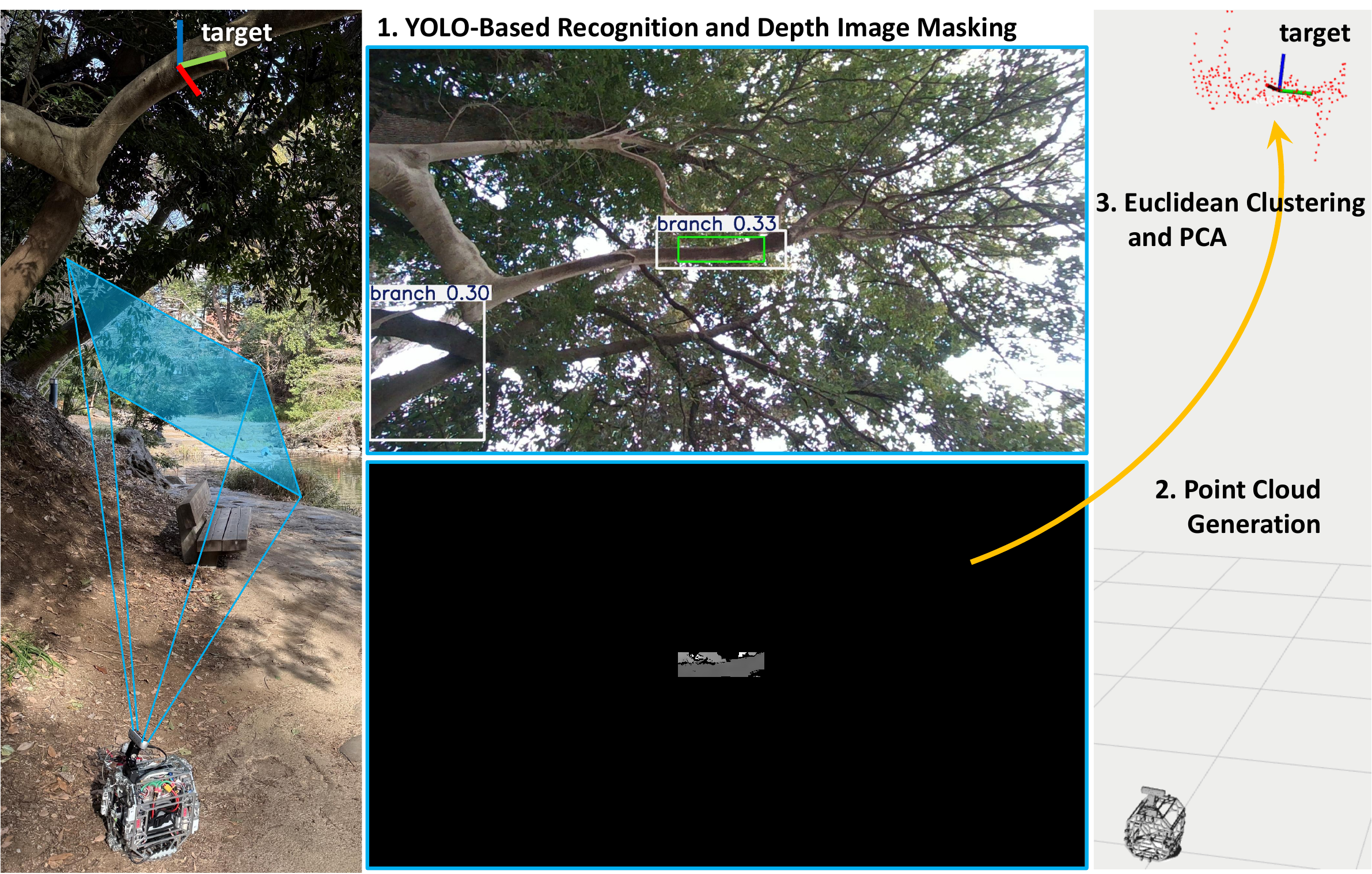}
      \vspace{-5.0ex}
      \caption{enviromental perceptionの概要．
      環境におけるワイヤ接続箇所をマスクした深度画像から点群を生成し，
      その点群からワイヤ接続対象の位置と向きを求める．
      }
      \vspace{-2.0ex}
      \label{fig:env_perception}
    \end{center}
  \end{figure}

  \begin{figure}[t]
    \begin{center}
      \includegraphics[width=1.0\columnwidth]{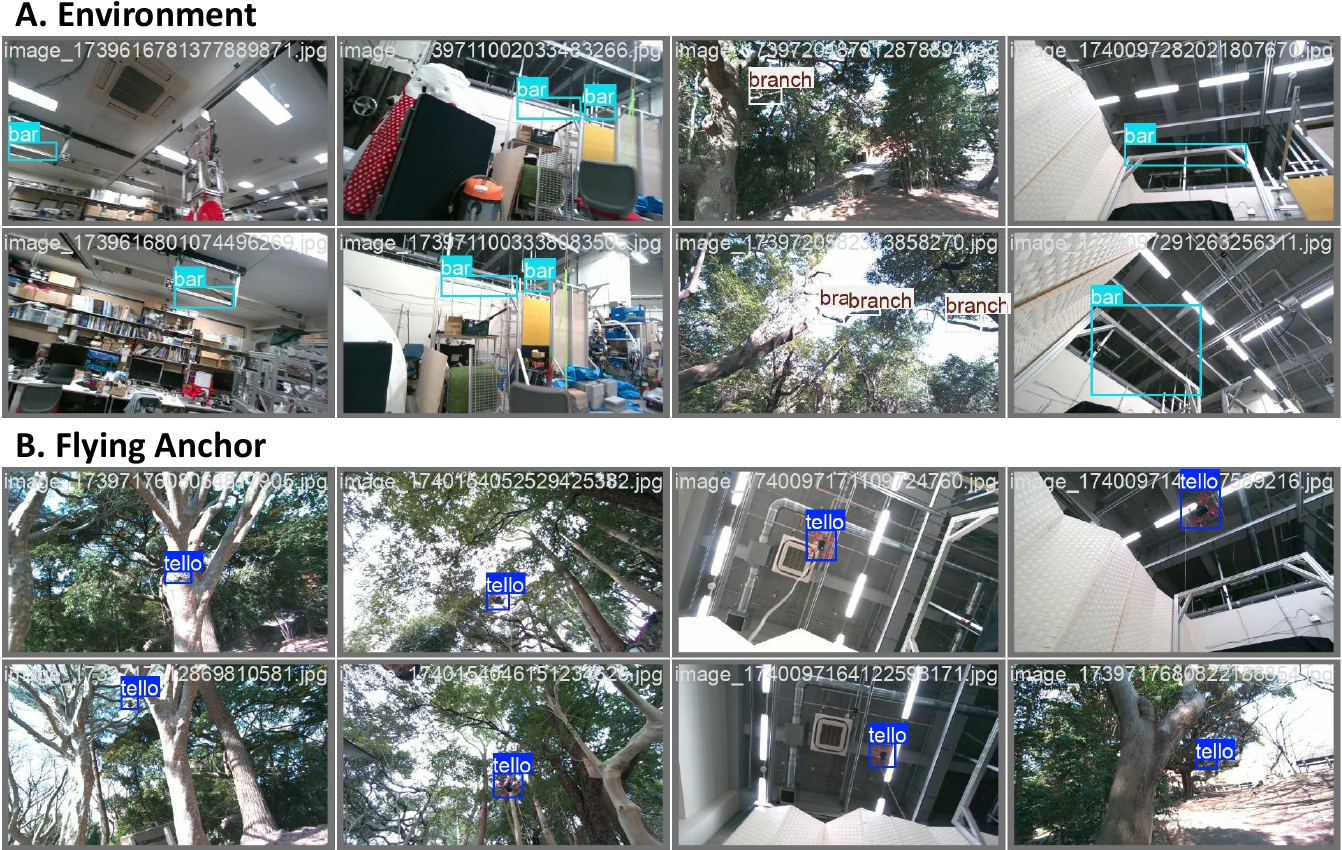}
      \vspace{-4.0ex}
      \caption{環境 (A) /飛行アンカー (B) 認識に用いた学習データの例．
      }
      \vspace{-5.0ex}
      \label{fig:annotation}
    \end{center}
  \end{figure}
  environmental perceptionの概要を\figref{fig:env_perception}に示す．

  はじめに，YOLO\cite{yolo} (YOLO11) により，
  CubiX上のカメラから得られたRGB画像からワイヤを接続できる場所のバウンディングボックスを得る (\figref{fig:env_perception}-1)．
  ここで用いるYOLOの推論モデルは，\figref{fig:annotation}Aに示される画像を一例として事前学習が行われている．
  事前学習では，本研究の実験で用いるフレームや，
  画像上で横向きに写りワイヤを接続できる太さの木の枝をCVAT\cite{boris_sekachev_2020_4009388}を用いて
  手動でアノテーションした画像を学習した．
  本研究ではフレームを示すbarと木の枝を示すbranchの2種類のラベルでワイヤを接続できる箇所を登録した．
  得られたバウンディングボックスを用いて，同時刻にカメラで得られた深度画像をマスクする．
  この際，バウンディングボックスより小さい範囲でマスクすることで，対象物体でない箇所の情報を減らしている．

  次に，マスクされた深度画像とカメラのパラメータを用いて3次元点群を構成する (\figref{fig:env_perception}-2)．

  そして，構成した点群からワイヤ接続箇所の位置，姿勢を求める (\figref{fig:env_perception}-3)．
  生成した点群にはノイズが含まれるため，ダウンサンプリング後にEuclidean clusteringで計算したクラスタから，
  最も近いクラスタを選択する．
  このクラスタの重心を求め，主成分分析\cite{jolliffe2002principal}により点群が分散している向きを求める．
  ここで，その重心を原点に持ち，主成分方向にY軸を持ち，鉛直平面上にZ軸を持ち，CubiXのカメラがある方向にX軸を持つ
  target座標系を計算する．
  世界座標系から見たtarget座標系の位置，姿勢をそれぞれ$\bm{p}^{i}_\text{target},\,\bm{q}^{i}_\text{target}$とおく．
  なお，バウンディングボックスの数だけこのtarget座標系が存在できるため，添字$i$を用いて区別する．
  点群のダウンサンプリング，クラスタリングには点群処理ライブラリであるPCL\cite{5980567}を用いた．
}%

\subsection{Targets Manager}
\switchlanguage%
{%
  \begin{figure}[t]
    \begin{center}
      \includegraphics[width=1.0\columnwidth]{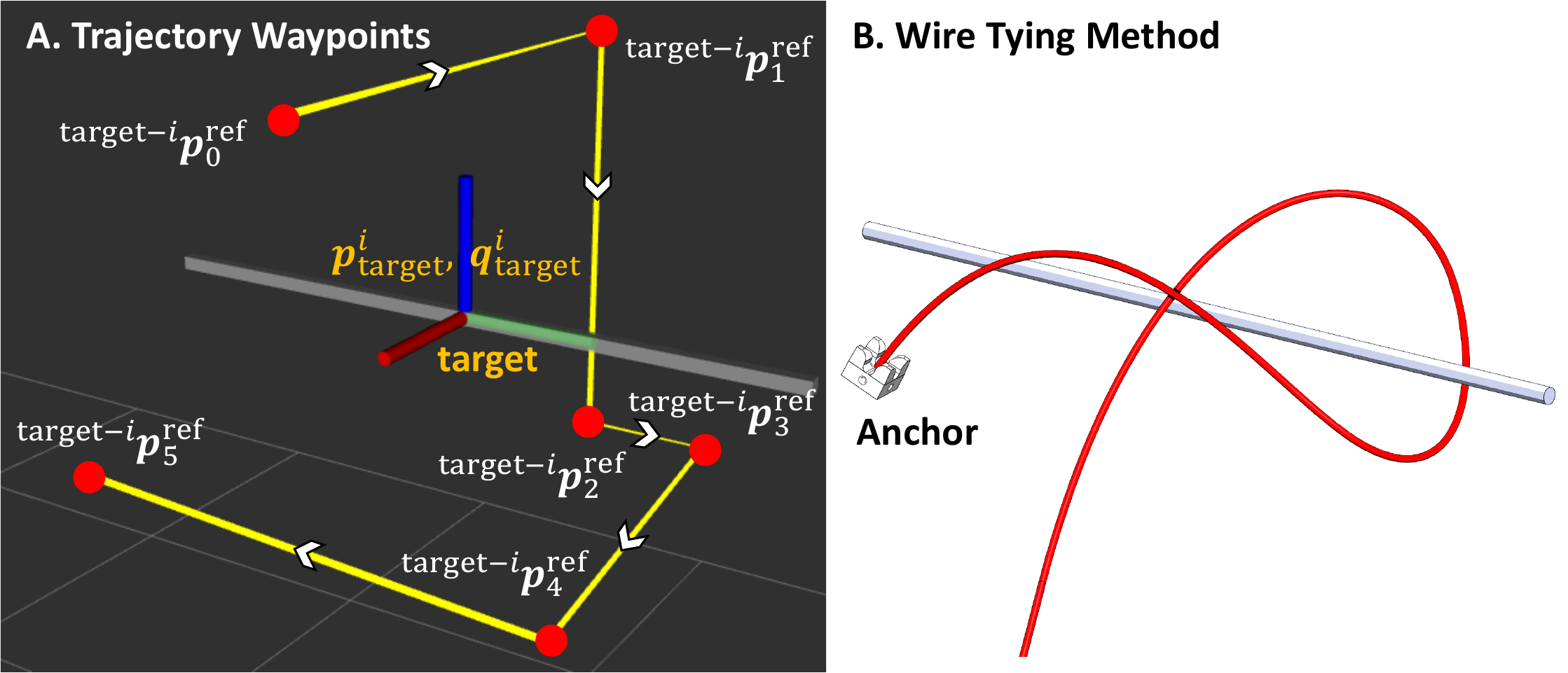}
      \vspace{-4.0ex}
      \caption{
        (A) The trajectory for attaching the wire to the environment, and (B) the method of wire tying. 
        The flying anchor follows this trajectory by position-controlling each waypoint.
      }
      \vspace{-5.0ex}
      \label{fig:traj_plan}
    \end{center}
  \end{figure}
  The multiple $\bm{p}^{i}_\text{target},\,\bm{q}^{i}_\text{target}$ obtained in Environmental Recognition 
  are managed using a Kalman filter\cite{1360855570047666048}. 
  The state and observations of the Kalman filter are the position and orientation of the \textit{target} coordinate frame. 
  When some $\bm{p}^{i}_\text{target},\,\bm{q}^{i}_\text{target}$ is observed, 
  if there is another \textit{target} coordinate whose position is within $c^\text{threshold}_\text{target}$, 
  that coordinate is updated; otherwise, a new entry is registered.

  Assuming that the target location does not change over time, 
  only the covariance matrix is updated 
  in the prediction step of the Kalman filter. 
  During the Kalman filter's update step, the state is corrected based on observation error, 
  because quaternions cannot be directly handled, 
  this study deals with the rotational vector $\bm{n}_e$ of the difference 
  between the observed orientation $\bm{q}_z$ and the state orientation $\bm{q}_x$.
  \begin{equation}
      \bm{n}_e = 2 \text{Im} (\bm{q}_z \cdot \overline{\bm{q}}_x)
  \end{equation}
  This uses the following relationship between the imaginary part of a quaternion $\text{Im}(\bm{q})$ 
  and the rotation vector $\bm{n}$. 
  \begin{equation}
      \text{Im}(\bm{q}) = \bm{n}\sin\cfrac{\theta}{2}
        \simeq \bm{n}\cfrac{\theta}{2}\quad
        \left(\sin\left|\cfrac{\theta}{2}\right| \ll 1\right)
  \end{equation}
  The state orientation is updated by computing the corresponding quaternion from $\bm{n}_e$.
}%
{%
  environmetal perceptionで得られた複数の$\bm{p}^{i}_\text{target},\,\bm{q}^{i}_\text{target}$を
  カルマンフィルタ\cite{1360855570047666048}を用いて管理する．
  カルマンフィルタの状態と観測はそれぞれtarget座標系の位置と姿勢である．
  ある$\bm{p}^{k}_\text{target},\,\bm{q}^{k}_\text{target}$が観測された時，
  それと位置が近い別の目標座標があれば，それを更新し，なければ新しく登録する．
  目標箇所は時間発展しないと仮定し，カルマンフィルタの予測ステップでは共分散行列のみがプロセスノイズの影響で更新される．
  カルマンフィルタの更新ステップでは，観測誤差に基づいて状態が補正されるが，そこでクォータニオンが直接扱えないため，
  本研究では，観測の姿勢$\bm{q}_z$と状態の姿勢$\bm{q}_x$の差分について，
  両者の差分の回転ベクトル$\bm{n}_e$を扱う．
  \begin{equation}
    \bm{n}_e = 2 \text{Im} (\bm{q}_z \cdot \overline{\bm{q}}_x)
  \end{equation}
  これは，クォータニオン$\bm{q}$の回転角$\theta$が$\text{sin}\left|\cfrac{\theta}{2}\right| \ll 1$を満たす場合における
  クォータニオンの実部$\text{Im}(\bm{q})$と回転ベクトル$\bm{n}$の以下の関係に利用している．
  \begin{equation}
    \text{Im}(\bm{q}) = \bm{n}\text{sin}\cfrac{\theta}{2}
    \simeq \bm{n}\cfrac{\theta}{2}
  \end{equation}
}%

\subsection{Trajectory Planning}
\switchlanguage%
{%
  A trajectory for attaching the wire to the environment is generated in the \textit{target} coordinate frame, 
  as shown in \figref{fig:traj_plan}A. 
  In order to satisfy the wire tying method required by this study's anchor (depicted in \figref{fig:traj_plan}B), 
  we predetermine waypoints $^{\text{target-}i}\bm{p}^\text{ref}_{j}$ for $j = 0, \dots, 5$.

  The flying anchor follows this path by repeatedly moving from its current position to each of the waypoints 
  $^{\text{target-}i}\bm{p}^\text{ref}_{j}$ in ascending order, starting with $j=0$. 
  Note that even if this trajectory is mirrored about the $XZ$-plane in the \textit{target} coordinate frame, 
  the wire can still be attached. 
}%
{%
  \begin{figure}[t]
    \begin{center}
      \includegraphics[width=1.0\columnwidth]{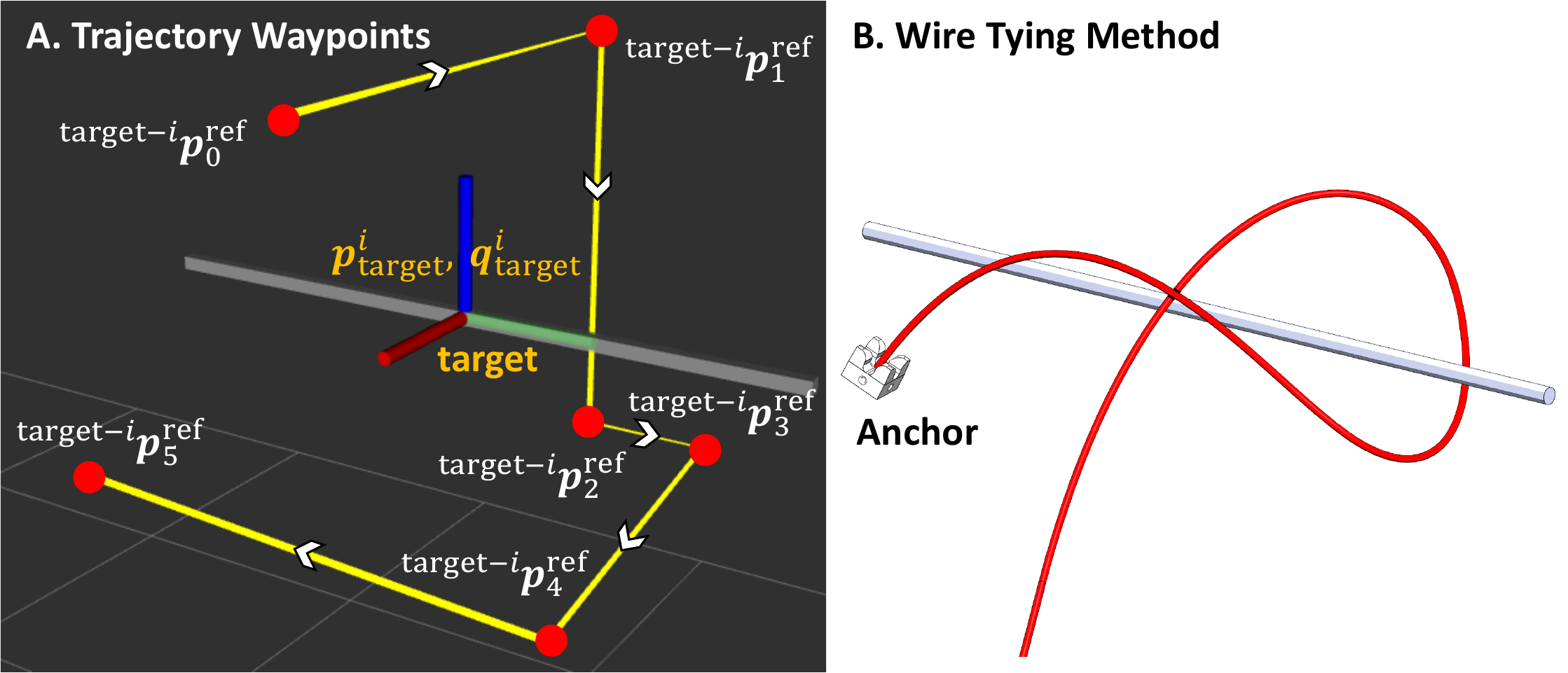}
      \vspace{-4.0ex}
      \caption{環境にワイヤを結ぶための軌跡 (A)と，ワイヤの結び方 (B)．
      飛行アンカーはこの軌跡に追従するために経由点に対して位置制御され，ワイヤを環境に結びつける．
      }
      \vspace{-5.0ex}
      \label{fig:traj_plan}
    \end{center}
  \end{figure}
  
  target座標系において，環境にワイヤを結びつけるための軌跡を\figref{fig:traj_plan}Aに示されるように生成する．
  \figref{fig:traj_plan}Bに示される，本研究のアンカーが要求するワイヤの巻き付け方が満たされるように，
  予め経由点$^\text{target}\bm{p}^\text{ref}_{i}$を決定する．
  本来，巻き付く環境の太さ，長さによってこの経路は変更する必要があるが，
  ある一定の太さ，長さの棒であることを仮定することで，本研究ではそれを陽に扱わないこととした．
  飛行アンカーは，経由点$^\text{target}\bm{p}^\text{ref}_{i}$について0から順番に現在地から向かうことを
  繰り返すことでこの経路へ追従することを実現する．
  このため，trajectory plannerは
  飛行アンカーの現在地に基づいて，向かうべき経由点$^\text{target}\bm{p}^\text{ref}_{i}$をflying anchor controllerへ渡す．
  なお，この経路をtarget座標系においてXZ平面について線対称なものにしてもワイヤを結ぶことができる．
}%

\subsection{Flying Anchor Recognition}
\switchlanguage%
{%
  \begin{figure}[t]
    \begin{center}
      \includegraphics[width=1.0\columnwidth]{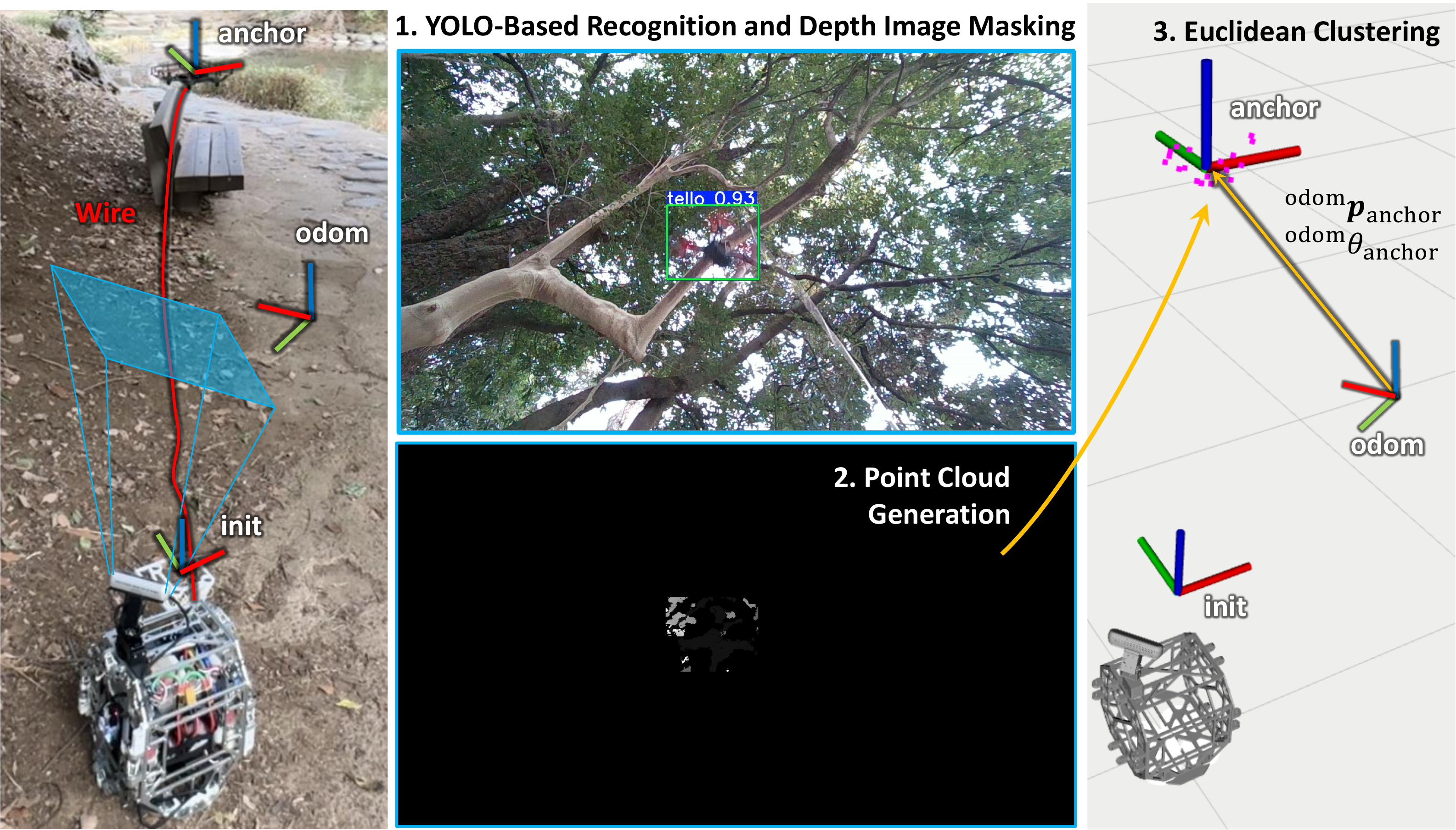}
      \vspace{-5.0ex}
      \caption{
        Overview of Flying Anchor Recognition. 
        A point cloud is generated from the depth image, masked at the flying anchor, 
        and the flying anchor's position is then determined.
      }
      \vspace{-6.0ex}
      \label{fig:anchor_perception}
    \end{center}
  \end{figure}

  \figref{fig:anchor_perception} shows an overview of Flying Anchor Recognition, 
  which largely follows the same procedure described in \subsecref{sec:env_recognition}. 
  First, bounding boxes for the flying anchors are obtained using YOLO11, 
  and these bounding boxes are used to mask the depth image (\figref{fig:anchor_perception}-1). 
  The YOLO model was pre-trained with images of flying anchors captured in our experimental environment. 
  Next, a point cloud is generated from the masked depth image (\figref{fig:anchor_perception}-2). 
  After downsampling this point cloud, we perform Euclidean clustering and select the nearest cluster. 
  The centroid of that cluster is computed as the flying anchor's estimated position $\hat{\bm{p}}_\text{anchor}$ 
  in the world coordinate system (\figref{fig:anchor_perception}-3).
}%
{%
  \begin{figure}[t]
    \begin{center}
      \includegraphics[width=1.0\columnwidth]{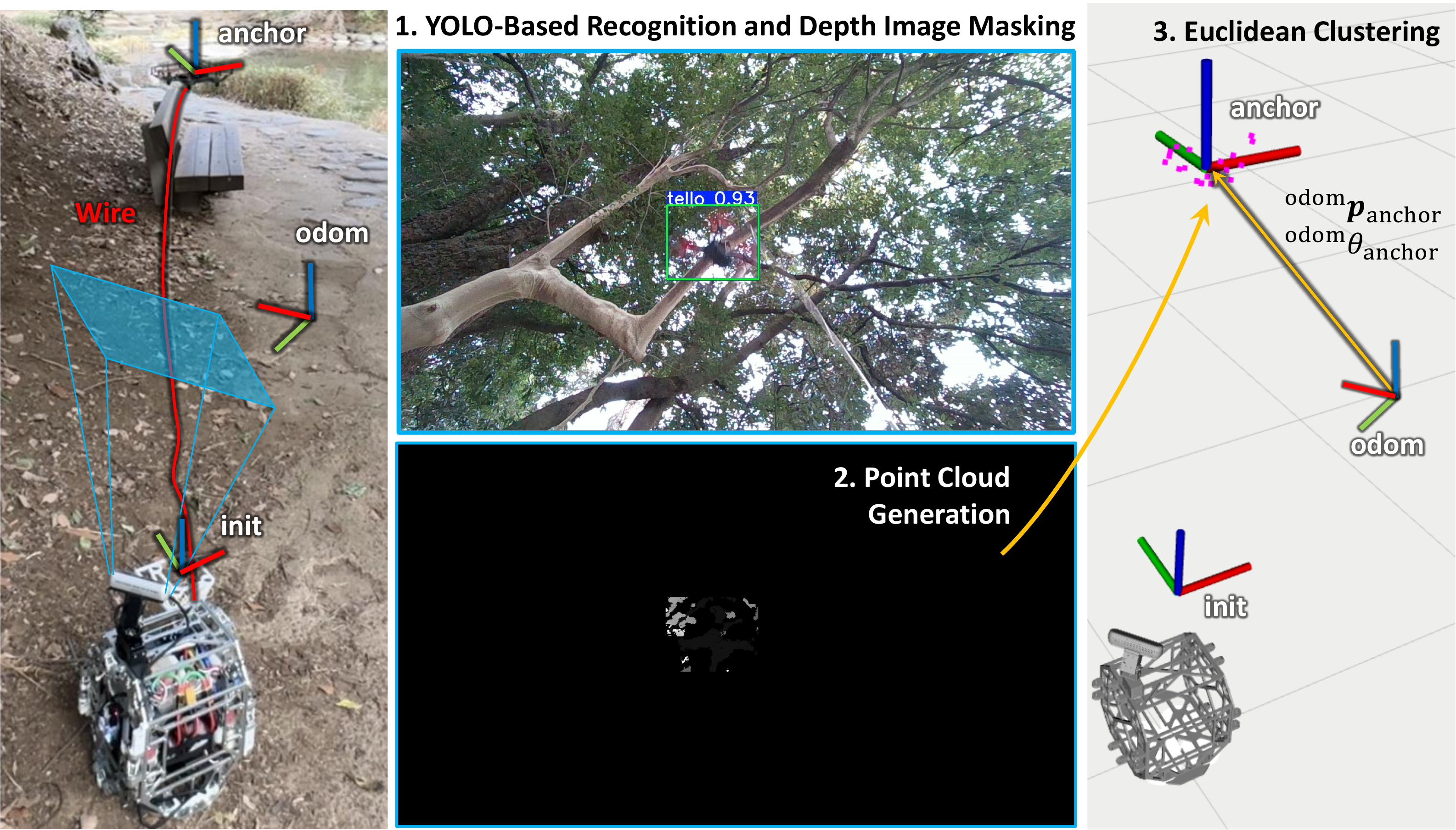}
      \vspace{-5.0ex}
      \caption{anchor perceptionの概要．
      飛行アンカーをマスクした深度画像から点群を生成し，
      その点群から飛行アンカーの位置を求める．
      }
      \vspace{-6.0ex}
      \label{fig:anchor_perception}
    \end{center}
  \end{figure}
  flying anchor perceptionの概要を\figref{fig:anchor_perception}に示す．

  はじめに，YOLO (YOLO11) により，
  CubiX上のカメラから得られたRGB画像から飛行アンカーのバウンディングボックスを得る (\figref{fig:anchor_perception}-1)．
  ここで用いるYOLOの推論モデルは，\figref{fig:annotation}Bに示される画像を一例として事前学習が行われている．
  事前学習では，本研究の実験環境でCubiXのカメラで撮影した飛行アンカーの画像を
  CVATを用いて手動でアノテーションしたものを学習した．
  得られたバウンディングボックスを用いて，同時刻にカメラで得られた深度画像をマスクする．
  次に，マスクされた深度画像とカメラのパラメータを用いて3次元点群を構成する (\figref{fig:anchor_perception}-2)．
  そして，構成した点群から飛行アンカーの位置を求める (\figref{fig:anchor_perception}-3)．
  生成した点群にはノイズが含まれるため，ダウンサンプリング後にEuclidean clusteringで計算したクラスタから，
  最も近いクラスタを選択する．
  このクラスタの重心を世界座標系での飛行アンカーの観測位置$\hat{\bm{p}}_\text{anchor}$として計算する．

}%

\subsection{Flying Anchor Control}
\switchlanguage%
{%
  First, the current position of the flying anchor is obtained by combining $\hat{\bm{p}}_\text{anchor}$, 
  which is provided by Flying Anchor Recognition, 
  with the information from the flying anchor's visual odometry using a Kalman filter. 
  The visual odometry computes the position $^\text{odom}\bm{p}_\text{anchor}$, 
  velocity $^\text{odom}\dot{\bm{p}}_\text{anchor}$, 
  and orientation $^\text{odom}\bm{q}_\text{anchor}$ of the flying anchor 
  relative to some uncertain \textit{odom} coordinate system, and transmits these values to CubiX.
  
  The state $\bm{x} \in \mathbb{R}^{11}$ and the observation $\bm{z} \in \mathbb{R}^{10}$ 
  for the Kalman filter are defined as follows:
  \begin{equation}
    \begin{split}
      \bm{x} &= \begin{bmatrix}
        \bm{p}_\text{anchor} & \dot{\bm{p}}_\text{anchor} & ^\text{anchor}\theta_\text{odom} & ^\text{odom}\bm{u} & ^\text{odom}\phi
      \end{bmatrix}^\top \\
      \bm{z} &= \begin{bmatrix}
        \hat{\bm{p}}_\text{anchor} & ^\text{odom}\bm{p}_\text{anchor} & ^\text{odom}\dot{\bm{p}}_\text{anchor} & ^\text{odom}\theta_\text{anchor}
      \end{bmatrix}^\top
    \end{split}
  \end{equation}
  Here, $\bm{u}$ and $\phi$ are the differences in position and yaw between the world coordinate system 
  and the \textit{odom} coordinate system, 
  and $\theta$ is the yaw angle of the flying anchor as viewed from the \textit{odom} coordinate system. 
  The state transition matrix $\bm{F}$ for the Kalman filter is defined as the following block matrix, where $\bm{I}$ is the identity matrix, 
  and $\Delta t$ is the timestep; it captures position changes due to velocity:
  \begin{equation}
    \bm{F} = 
    \begin{bmatrix}
      \begin{array}{cc} 
        \bm{I}_3 & \begin{array}{cc} \Delta t \bm{I}_3 & \bm{0}_{3 \times 5} \end{array} \\ 
        \bm{0}_{8 \times 3} & \bm{I}_8
      \end{array}
    \end{bmatrix}
  \end{equation}
  Assuming the flying anchor's roll and pitch angles are sufficiently close to zero, 
  the observation can be written as a function $\bm{h}(\bm{x})$ of the state:
  \begin{equation}
    \bm{z} = \bm{h}(\bm{x}) = \begin{bmatrix} 
      \bm{p}_\text{anchor} \\
      ^\text{odom}\bm{T}(^\text{odom}\phi)\bm{p}_\text{anchor} + ^\text{odom}\bm{u} \\
      ^\text{odom}\bm{T}(^\text{odom}\phi)\dot{\bm{p}}_\text{anchor} \\
      -^\text{anchor}\theta_\text{odom}
    \end{bmatrix}
  \end{equation}
  where $\bm{T}$ is a rotation matrix. 
  We define $\bm{H}$ as the Jacobian of $\bm{h}(\bm{x})$ with respect to $\bm{x}$. 
  Using these definitions, the prediction step of the Kalman filter is given by:
  \begin{equation}
    \begin{split}
      \bm{x} &\leftarrow \bm{F} \bm{x} \\
      \bm{P} &\leftarrow \bm{F} \bm{P} \bm{F}^T + \bm{G} \bm{Q} \bm{G}^\top
    \end{split}
  \end{equation}
  and the update step is:
  \begin{equation}
    \begin{split}
      \bm{e} &= \bm{z} - \bm{h}(\bm{x}) \\
      \bm{S} &= \bm{R} + \bm{H}\bm{P}\bm{H}^\top \\
      \bm{K} &= \bm{P}\bm{H}^\top\bm{S}^{-1} \\
      \bm{x} &\leftarrow \bm{x} + \bm{K}\bm{e} \\
      \bm{P} &\leftarrow \bm{I}_{11} - \bm{K}\bm{H}\bm{P}
    \end{split}
  \end{equation}
  Here, $\bm{G}$, $\bm{Q}$, and $\bm{R}$ are the process noise influence matrix, 
  the process noise covariance matrix, and the observation noise covariance matrix, respectively.
  
  We also define an \textit{init} coordinate frame on the helipad where the flying anchor takes off. 
  The observation $\hat{\bm{p}}_\text{anchor}$ is incorporated into the update step if, at the first time step, 
  $\hat{\bm{p}}_\text{anchor}$ is within $c^\text{threshold}_\text{anchor}$ of the $z$-axis of the \textit{init} coordinate frame, 
  and thereafter if it is within $c^\text{threshold}_\text{anchor}$ of $\bm{p}_\text{anchor}$.
  This makes it possible to observe multiple flying anchors simultaneously.
  
  Finally, the computed current position $\bm{p}_\text{anchor}$ is brought closer 
  to the waypoint $^{\text{target-}i}\bm{p}^\text{ref}_{j}$ from the Trajectory Planner by applying PID control.
  \begin{equation}
    C_x, C_y, C_z = \text{PID}\left(\bm{p}_\text{anchor} \to ^{\text{target-}i}\bm{p}^\text{ref}_{j}, \dot{\bm{p}}_\text{anchor} \to \bm{0}\right)
  \end{equation}
  By transmitting this computed control input to the flying anchor, 
  the system achieves the wire-attachment motion in the environment.
}%
{%
  まず，飛行アンカーの現在位置をflying anchor perceptionで得られた$\hat{\bm{p}}_\text{anchor}$と，
  飛行アンカーが行うビジュアルオドメトリによる情報とをカルマンフィルタを用いて合わせて求める．
  ビジュアルオドメトリでは，
  ある不確定のodom座標系からみた飛行アンカーの位置$^\text{odom}\bm{p}_\text{anchor}$，
  速度$^\text{odom}\dot{\bm{p}}_\text{anchor}$，姿勢$^\text{odom}\bm{q}_\text{anchor}$が計算され，CubiXに送信される．
  カルマンフィルタの状態$\bm{x}_\text{anchor} \in \mathbb{R}^{11}$と観測$\bm{z}_\text{anchor} \in \mathbb{R}^{10}$は以下のように定義される．
  \begin{equation}
    \begin{split}
      \bm{x}_\text{anchor} &= \begin{bmatrix} \bm{p}_\text{anchor} & \dot{\bm{p}}_\text{anchor} & ^\text{anchor}\theta_\text{odom} & ^\text{odom}\bm{u} & ^\text{odom}\phi \end{bmatrix}^\top \\
      \bm{z}_\text{anchor} &= \begin{bmatrix} \hat{\bm{p}}_\text{anchor} & ^\text{odom}\bm{p}_\text{anchor} & ^\text{odom}\dot{\bm{p}}_\text{anchor} & ^\text{odom}\theta_\text{anchor}\end{bmatrix}^\top
    \end{split}
  \end{equation}
  ここで，$\bm{u}, \phi$は世界座標系とodom座標系との位置，yaw角の差分であり，
  $\theta$はodom座標系からみた飛行アンカーのyaw角である．
  状態遷移行列 $\bm{F}$ は、$\bm{I}$を単位行列，$\Delta t$をステップ時間として，
  速度による位置の時間発展を考えることで次のブロック行列で定義される。
  \begin{equation}
    \bm{F} = 
    \begin{bmatrix}
      \begin{array}{cc} 
        {\bm{I}_3 & \begin{array}{cc} \Delta t \bm{I}_3 & \bm{0}_{3 \times 5} \end{array}} \\ 
        \bm{0}_{8 \times 3} & \bm{I}_8
      \end{array}
    \end{bmatrix}
  \end{equation}
  飛行アンカーのroll，pitch角は十分0に近いとして，観測を状態の関数$h(\bm{x}_\text{anchor})$で表すと，
  \begin{equation}
    \bm{z}_\text{anchor} = \bm{h}(\bm{x}_\text{anchor}) = \begin{bmatrix} 
      \bm{p}_\text{anchor} \\
      ^\text{odom}\bm{T}(^\text{odom}\phi) + ^\text{odom}\bm{u} \\
      ^\text{odom}\bm{T}(^\text{odom}\phi)\dot{\bm{p}}_\text{anchor} \\
      -^\text{anchor}\theta_\text{odom}
    \end{bmatrix}
  \end{equation}
  なお，$\bm{T}$は回転行列を表す．
  この$\bm{h}(\bm{x}_\text{anchor})$を$\bm{x}_\text{anchor}$で偏微分した行列を$\bm{H}$と定義する．
  これらを用いて，カルマンフィルタの予測ステップは以下のように書ける．
  \begin{equation}
    \begin{split}
      \bm{x}_\text{anchor} &\leftarrow \bm{F} \bm{x}_\text{anchor} \\
      \bm{P}_\text{anchor} &\leftarrow \bm{F} \bm{P}_\text{anchor} \bm{F}^T + \bm{G}_\text{anchor} \bm{Q}_\text{anchor} \bm{G}_\text{anchor}^\top
    \end{split}
  \end{equation}
  また，カルマンフィルタの更新ステップは以下のように書ける．
  \begin{equation}
    \begin{split}
      \bm{e}_\text{anchor} &= \bm{z}_\text{anchor} - \bm{h}(\bm{x}_\text{anchor}) \\
      \bm{S}_\text{anchor} &= \bm{R}_\text{anchor} + \bm{H}\bm{P}_\text{anchor}\bm{H}^\top \\
      \bm{K}_\text{anchor} &= \bm{P}\bm{H}^\top\bm{S}^{-1}_\text{anchor} \\
      \bm{x}_\text{anchor} &\leftarrow \bm{x}_\text{anchor} + \bm{K}_\text{anchor}\bm{e}_\text{anchor} \\
      \bm{P}_\text{anchor} &\leftarrow \bm{I}_{11} - \bm{K}_\text{achor}\bm{H}\bm{P}_\text{anchor}
    \end{split}
  \end{equation}
  なお、$\bm{G}_{\text{anchor}}, \bm{Q}_{\text{anchor}}, \bm{R}_{\text{anchor}}$ は、
  それぞれプロセスノイズの影響行列、プロセスノイズの共分散行列、および観測ノイズの共分散行列である。
  また，飛行アンカーが離陸するhelipadにinit座標系を定義する．
  観測$\hat{\bm{p}}_\text{anchor}$による更新ステップは，その位置$\hat{\bm{p}}_\text{anchor}$が，
  初回はinit座標系のz軸に一定以上近い場合，
  初回以降は$\bm{p}_\text{anchor}$にある一定以上近い場合に行われる．
  それにより，複数の飛行アンカーの観測を同時に行うことができる．

  そして，計算した現在位置$\bm{p}_\text{anchor}$を
  trajectory plannerから渡される$^\text{target}\bm{p}^\text{ref}_{i}$へ近づけるためにPID制御を行う．
  \begin{equation}
    C_x, C_y, C_z = \text{PID}\left(\bm{p}_\text{anchor} \to ^\text{target}\bm{p}^\text{ref}_{i}, \dot{\bm{p}}_\text{anchor} \to \bm{0}\right)
  \end{equation}
  このように計算した操作量を飛行アンカーへ送信することで，ワイヤを環境に結びつける動作を実現する．
}%

\section{Experiments} \label{sec:experiments}
\subsection{Cliff Climbing Experiment}
\switchlanguage%
{%
  \begin{figure}[t]
    \begin{center}
      \includegraphics[width=1.0\columnwidth]{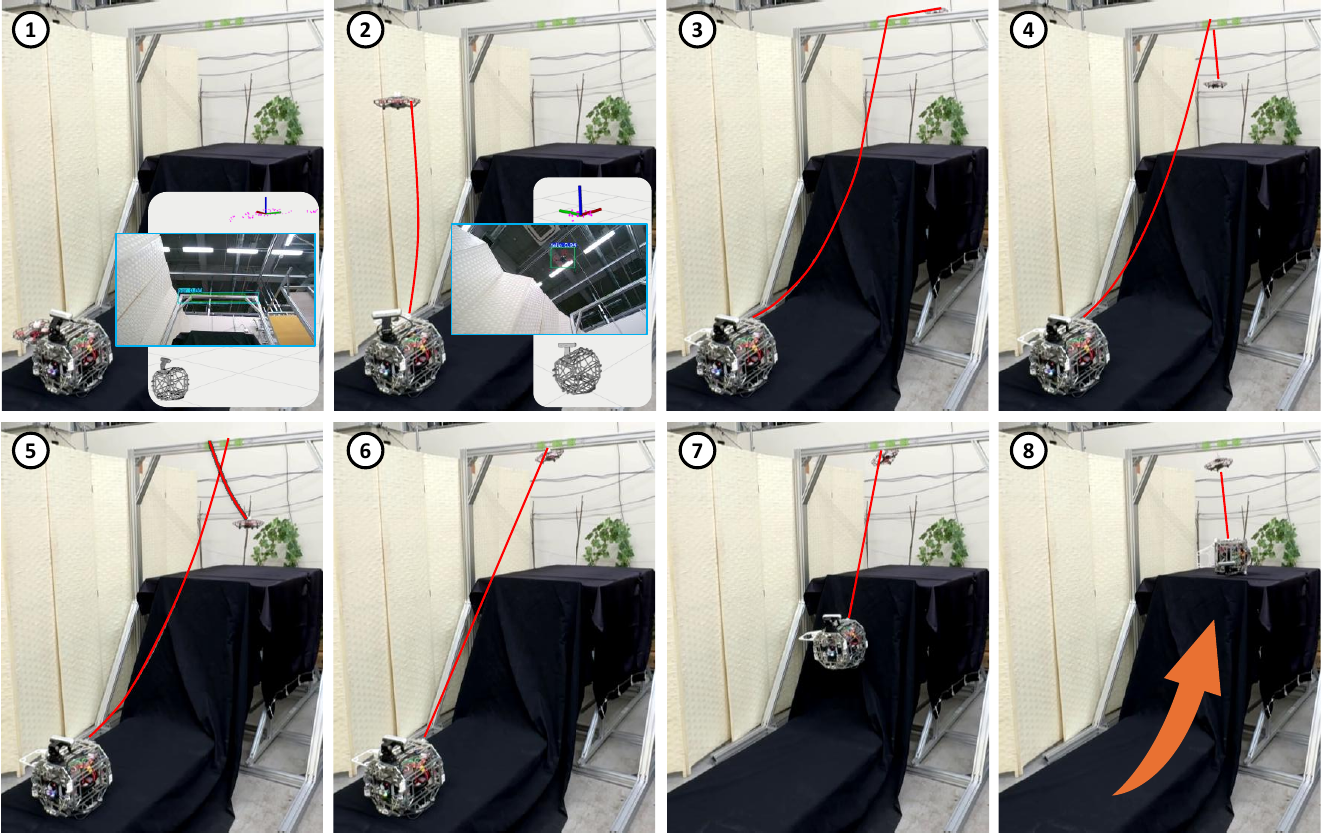}
      \vspace{-5.0ex}
      \caption{
        Scenes from the cliff climbing experiment, showing CubiX and the flying anchor. 
        CubiX attaches a wire to the bar and climbs the cliff by winding that wire.
      }
      \vspace{-3.0ex}
      \label{fig:exp1_pic}
    \end{center}
  \end{figure}

  \begin{figure}[t]
    \begin{center}
      \includegraphics[width=0.6\columnwidth]{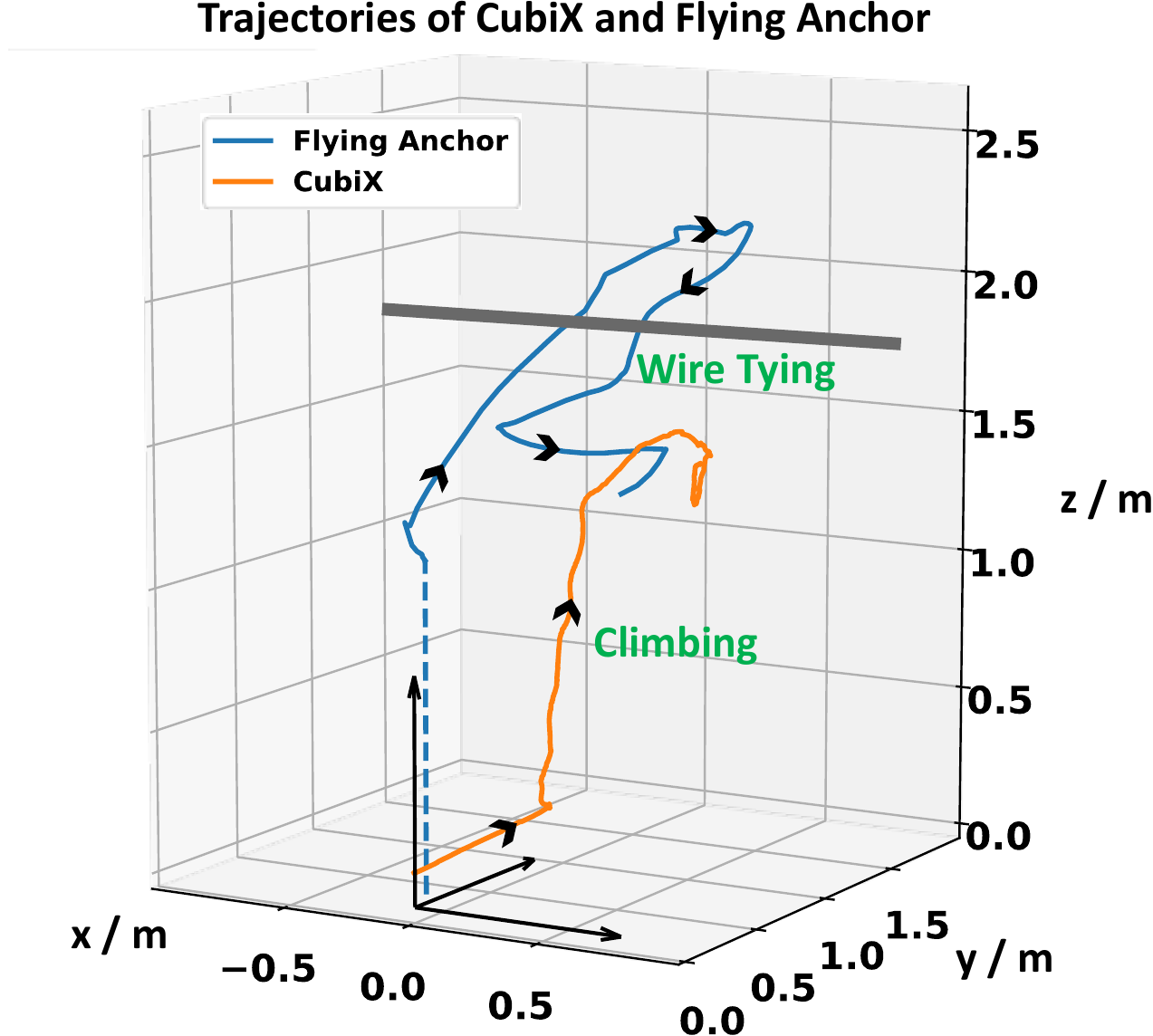}
      \vspace{-2.0ex}
      \caption{
        Trajectories of CubiX and the flying anchor in the cliff climbing experiment. 
        The anchor flies around the recognized bar to attach the wire, enabling CubiX to climb the cliff.
      }
      \vspace{-5.0ex}
      \label{fig:exp1_data}
    \end{center}
  \end{figure}

  To evaluate the functionality of the proposed system, 
  we conducted an experiment in which a single wire was attached to a bar in the environment, 
  and CubiX climbed a cliff by winding the wire. 
  \figref{fig:exp1_pic} shows CubiX attaching the wire to the environment and climbing the cliff via wire driving, 
  and \figref{fig:exp1_data} shows the trajectories of CubiX and the flying anchor during this process. 

  In \figref{fig:exp1_pic}\ctext{1}, 
  CubiX recognizes the bar in the environment using its camera and computes its position and orientation. 
  In \figref{fig:exp1_pic}\ctext{2}, after the flying anchor has taken off, 
  CubiX recognizes the flying anchor via its camera and acquires the anchor's position. 
  In \figref{fig:exp1_pic}\ctext{3}--\ctext{5}, the flying anchor follows a generated path around the recognized bar, 
  thus attaching the wire to the bar. 
  In \figref{fig:exp1_pic}\ctext{6}, CubiX applies tension to the wire, 
  thereby securing it via the flying anchor.
  In \figref{fig:exp1_pic}\ctext{7}--\ctext{8}, CubiX increases the tension on the wire and climbs the cliff. 
  This demonstrates that the proposed system functions properly, 
  allowing CubiX to autonomously attach its own wire to the environment and drive itself using that wire.

}%
{%
  \begin{figure}[t]
    \begin{center}
      \includegraphics[width=1.0\columnwidth]{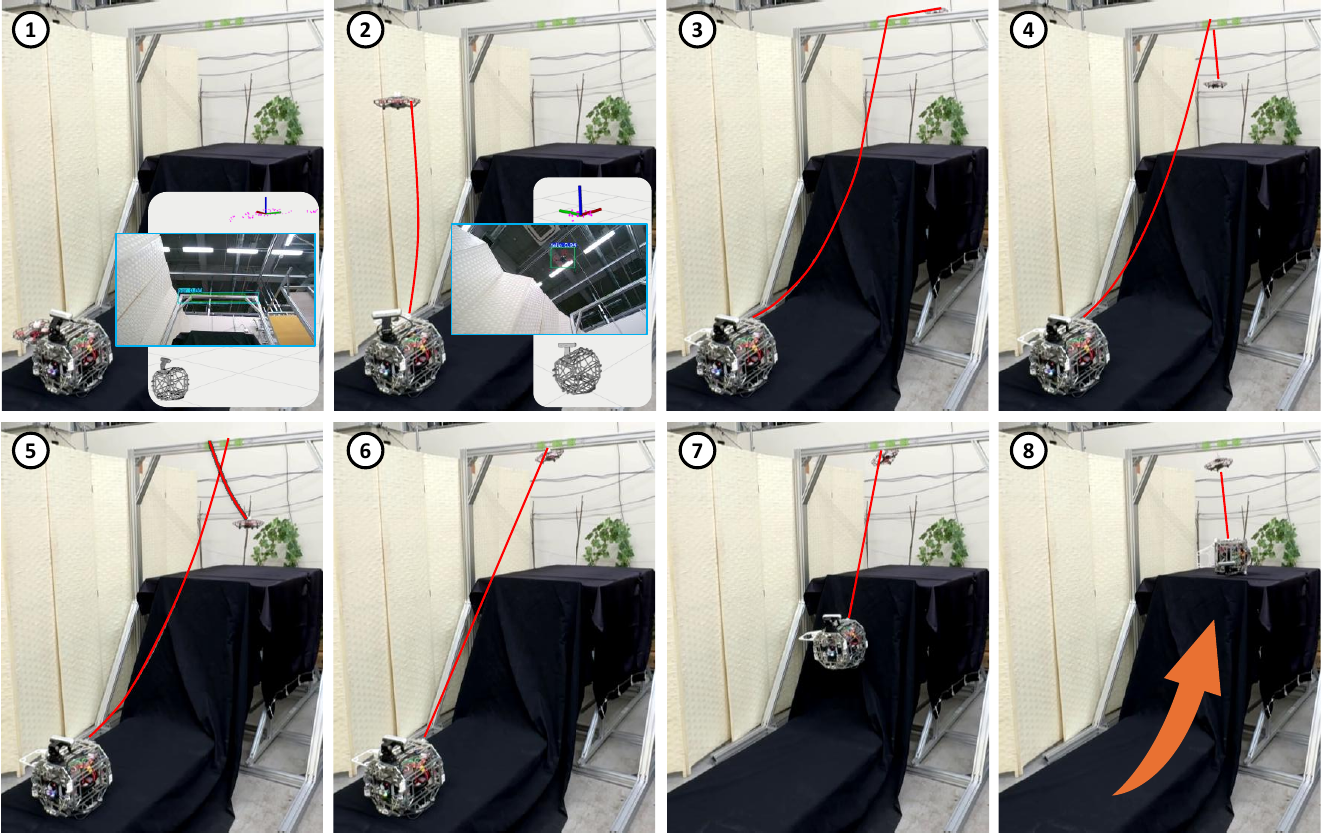}
      \vspace{-5.0ex}
      \caption{崖上り実験におけるCubiXと飛行アンカーの様子．
      CubiXがバーにワイヤを接続し，そのワイヤを巻き取って駆動し崖を登る．
      }
      \vspace{-5.0ex}
      \label{fig:exp1_pic}
    \end{center}
  \end{figure}

  \begin{figure}[t]
    \begin{center}
      \includegraphics[width=0.8\columnwidth]{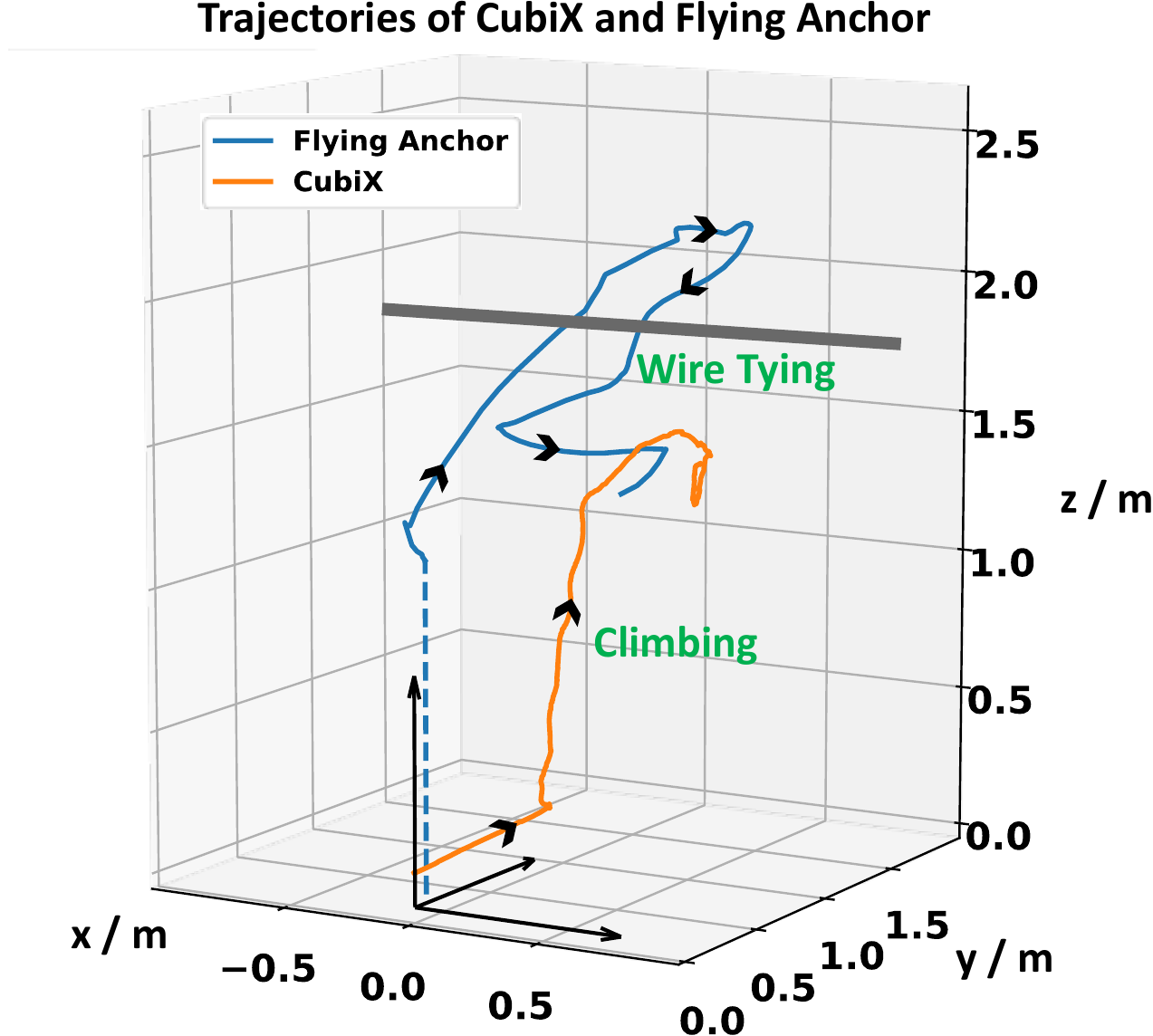}
      \vspace{-2.0ex}
      \caption{崖上り実験におけるCubiXと飛行アンカーの軌跡．
      認識したバーの周辺を飛行してワイヤを巻き付け，それを用いてCubiXが崖を登ったことがわかる．
      }
      \vspace{-3.0ex}
      \label{fig:exp1_data}
    \end{center}
  \end{figure}

  本研究で提案するシステムが機能するかを評価するため，
  1本のワイヤを環境であるbarに接続し，それを巻き取って駆動することによって崖を登る実験を行った．
  CubiXが環境にワイヤを接続し，ワイヤ駆動によって崖を登る様子を\figref{fig:exp1_pic}に，
  その際のCubiXと飛行アンカーの軌跡を\figref{fig:exp1_data}に示す．
  \figref{fig:exp1_pic}\ctext{1}で，環境に存在するバーをCubiXのカメラで認識し，その位置と方向を計算した．
  \figref{fig:exp1_pic}\ctext{2}で，飛行アンカーを離陸させた後，飛行アンカーをCubiXのカメラから認識し，
  飛行アンカーの世界座標系の位置を取得した．
  \figref{fig:exp1_pic}\ctext{3}--\ctext{5}で，認識したバーの周りに生成した経路を飛行アンカーが追従することで，
  ワイヤをバーに結びつけた．
  \figref{fig:exp1_pic}\ctext{6}で，CubiXがワイヤに張力をかけることで，バーに結ばれたワイヤを飛行アンカーで固定した．
  \figref{fig:exp1_pic}\ctext{7}--\ctext{8}で，ワイヤにかけた張力を増やすことで，CubiXが崖を登った．
  これにより，本研究のシステムが機能し，CubiXが自身のワイヤを環境に自律的に接続して，
  そのワイヤを用いて駆動できるということが示された．
}%

\subsection{Outdoor Experiment}
\switchlanguage%
{%
  \begin{figure}[t]
    \begin{center}
      \includegraphics[width=1.0\columnwidth]{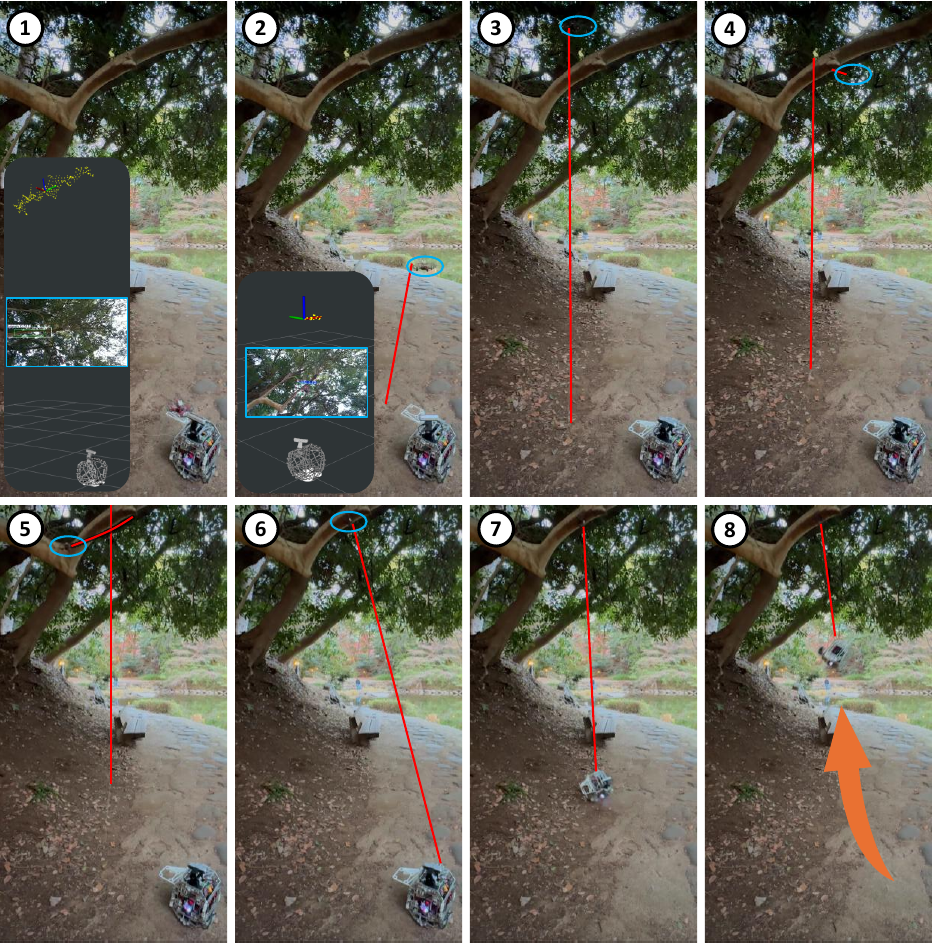}
      \vspace{-5.0ex}
      \caption{
        Scenes from the outdoor experiment, showing CubiX and the flying anchor. 
        CubiX recognizes a tree branch, attaches the wire, and drives using that wire.
      }
      \vspace{-3.0ex}
      \label{fig:exp2_pic}
    \end{center}
  \end{figure}

  \begin{figure}[t]
    \begin{center}
      \includegraphics[width=0.6\columnwidth]{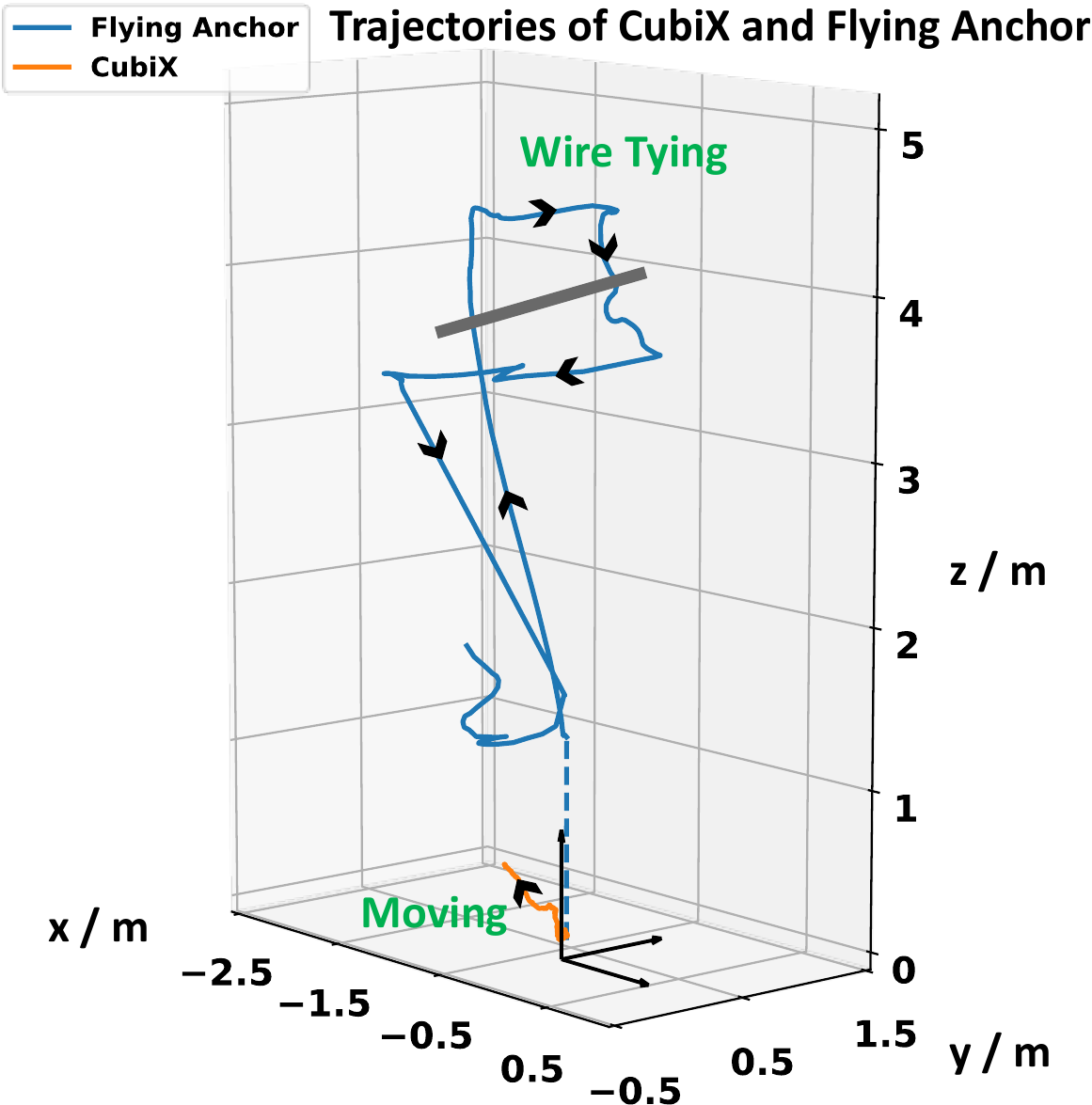}
      \vspace{-2.0ex}
      \caption{
        Trajectories of CubiX and the flying anchor in the outdoor experiment. 
        The flying anchor flies around the recognized branch to attach the wire, enabling CubiX to drive.
      }
      \vspace{-5.0ex}
      \label{fig:exp2_data}
    \end{center}
  \end{figure}

  We also evaluated whether the proposed system can attach wires to environments that are not specially prepared, 
  conducting an experiment in which a single wire was attached to a tree branch in an outdoor setting, 
  and CubiX was driven by winding that wire. 
  \figref{fig:exp2_pic} shows CubiX connecting the wire to the environment and then driving with that wire, 
  and \figref{fig:exp2_data} shows the trajectories of CubiX and the flying anchor. 
  During wire driving in this experiment, 
  CubiX swung back and forth
  , causing computational failures in the V-SLAM camera. 
  As a result, CubiX's trajectory data terminate when it rises significantly into the air.

  In \figref{fig:exp2_pic}\ctext{1}, 
  CubiX recognizes the branch in the environment using its camera and computes its position and orientation. 
  In \figref{fig:exp2_pic}\ctext{2}, after the flying anchor has taken off, 
  CubiX recognizes the flying anchor via its camera and acquires the anchor's position. 
  In \figref{fig:exp2_pic}\ctext{3}--\ctext{5}, the flying anchor follows a generated path around the recognized branch, 
  thus attaching the wire to the tree. 
  In \figref{fig:exp2_pic}\ctext{6}, CubiX applies tension to the wire, 
  thereby securing it via the flying anchor.
  In \figref{fig:exp2_pic}\ctext{7}--\ctext{8}, CubiX drives by increasing the wire tension and moves into the air. 
  This result demonstrates that the proposed system also functions in outdoor environments that have not been specially prepared, 
  enabling CubiX to perform wire driving.

  However, in \figref{fig:exp2_pic}\ctext{5}, 
  the flying anchor collides with a different branch near the final waypoint of its path, 
  although it was still able to attach the wire successfully. 
  Ideally, after recognizing the attachment point, 
  the system should identify obstacles in the surrounding area and generate a path that avoids them.
}%
{%
  \begin{figure}[t]
    \begin{center}
      \includegraphics[width=1.0\columnwidth]{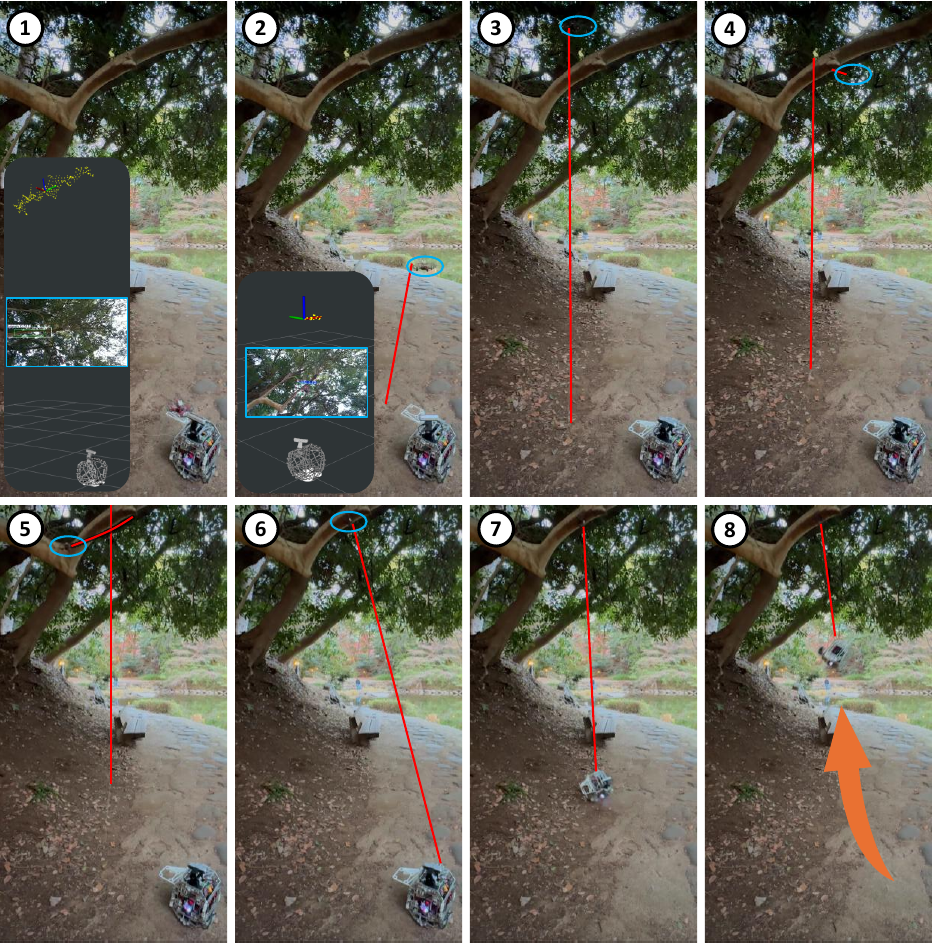}
      \vspace{-5.0ex}
      \caption{屋外実験におけるCubiXと飛行アンカーの様子．
      CubiXが木の枝を認識してワイヤを接続し，そのワイヤを用いて駆動する．
      }
      \vspace{-5.0ex}
      \label{fig:exp2_pic}
    \end{center}
  \end{figure}

  \begin{figure}[t]
    \begin{center}
      \includegraphics[width=0.8\columnwidth]{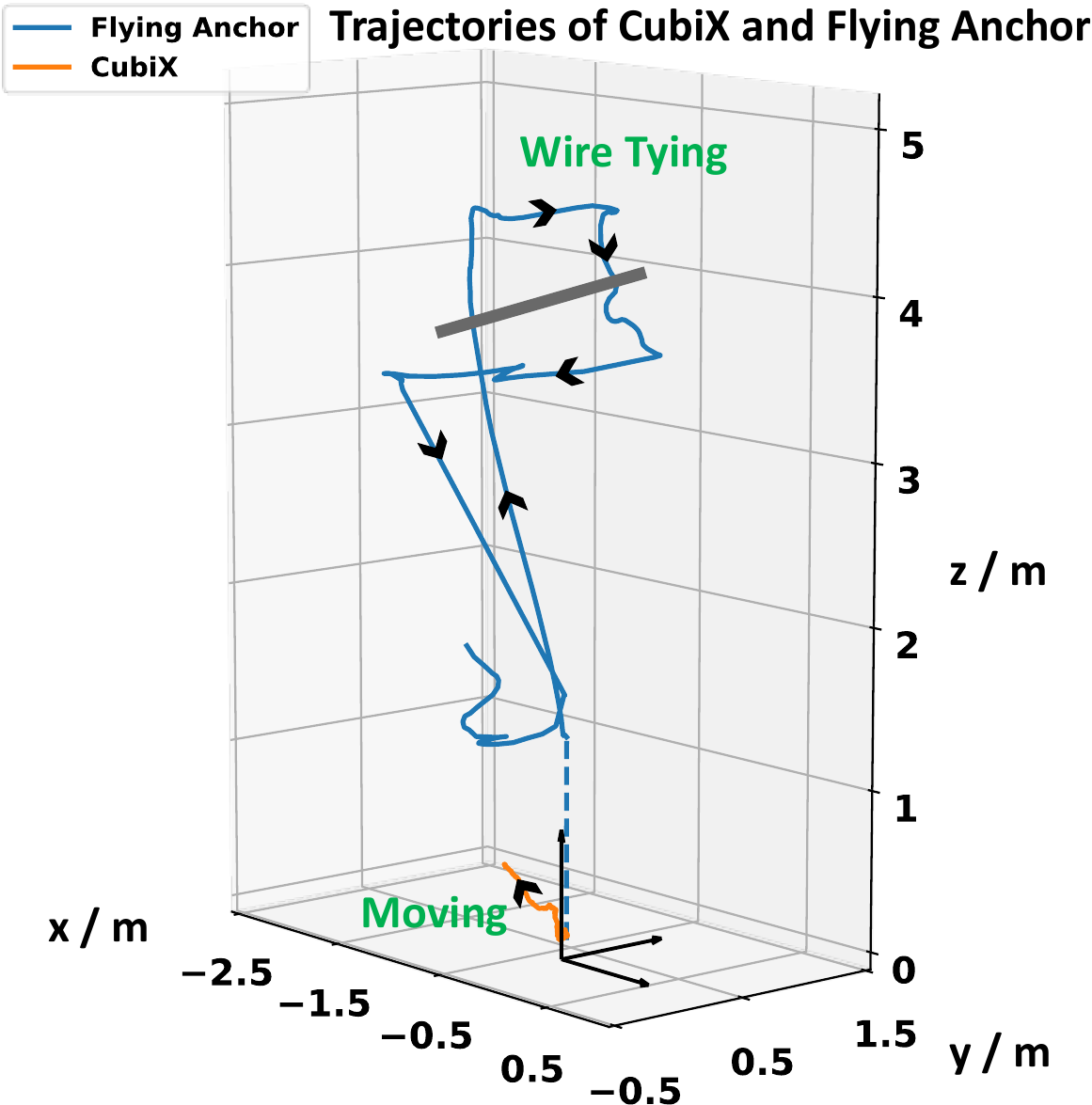}
      \vspace{-2.0ex}
      \caption{屋外実験におけるCubiXと飛行アンカーの軌跡．
      認識した木の周辺を飛行してワイヤを巻き付け，それを用いてCubiXが駆動したことがわかる．
      }
      \vspace{-5.0ex}
      \label{fig:exp2_data}
    \end{center}
  \end{figure}

  特別に整備されていない屋外環境についても，
  本研究のシステムを用いて環境にワイヤを接続できるかを評価する．
  1本のワイヤを環境である木の枝に接続し，それを巻き取って駆動する実験を行った．
  CubiXが環境にワイヤを接続し，そのワイヤで駆動する様子を\figref{fig:exp2_pic}に，
  その際のCubiXと飛行アンカーの軌跡を\figref{fig:exp2_data}に示す．
  なお，本実験のワイヤ駆動ではCubiXが振り子のように大きく振動し，CubiXに搭載されているV-SLAMカメラが計算不良を起こしたため，
  CubiXの軌跡データは，CubiXが空中へ大きく上昇するところで途切れている．
  \figref{fig:exp2_pic}\ctext{1}で，環境に存在する枝をCubiXのカメラで認識し，その位置と方向を計算した．
  \figref{fig:exp2_pic}\ctext{2}で，飛行アンカーを離陸させた後，飛行アンカーをCubiXのカメラから認識し，
  飛行アンカーの世界座標系の位置を取得した．
  \figref{fig:exp2_pic}\ctext{3}--\ctext{5}で，認識した枝の周りに生成した経路を飛行アンカーが追従することで，
  ワイヤを木に結びつけた．
  \figref{fig:exp2_pic}\ctext{6}で，CubiXがワイヤに張力をかけることで，木に結ばれたワイヤを飛行アンカーで固定した．
  \figref{fig:exp2_pic}\ctext{7}--\ctext{8}で，ワイヤにかけた張力を増やすことで駆動し，CubiXが空中へ移動した．
  これにより，本研究のシステムが特別に整備されていない屋外環境についても機能し，
  CubiXがワイヤ駆動を形成できることが示された．

  しかし，\figref{fig:exp2_pic}\ctext{5}で，
  目標軌跡終点付近にて， 飛行アンカーがワイヤを結びつけている枝とは別の枝に衝突している．
  それでも問題なくワイヤを結び付けられていたが，
  本来はワイヤを接続する箇所を認識した後，その周辺の障害物を確認し，それを回避するような軌跡を計算できる必要がある．
}%

\subsection{Multiple-Wire Experiment}
\switchlanguage%
{%
  \begin{figure}[t]
    \begin{center}
      \includegraphics[width=1.0\columnwidth]{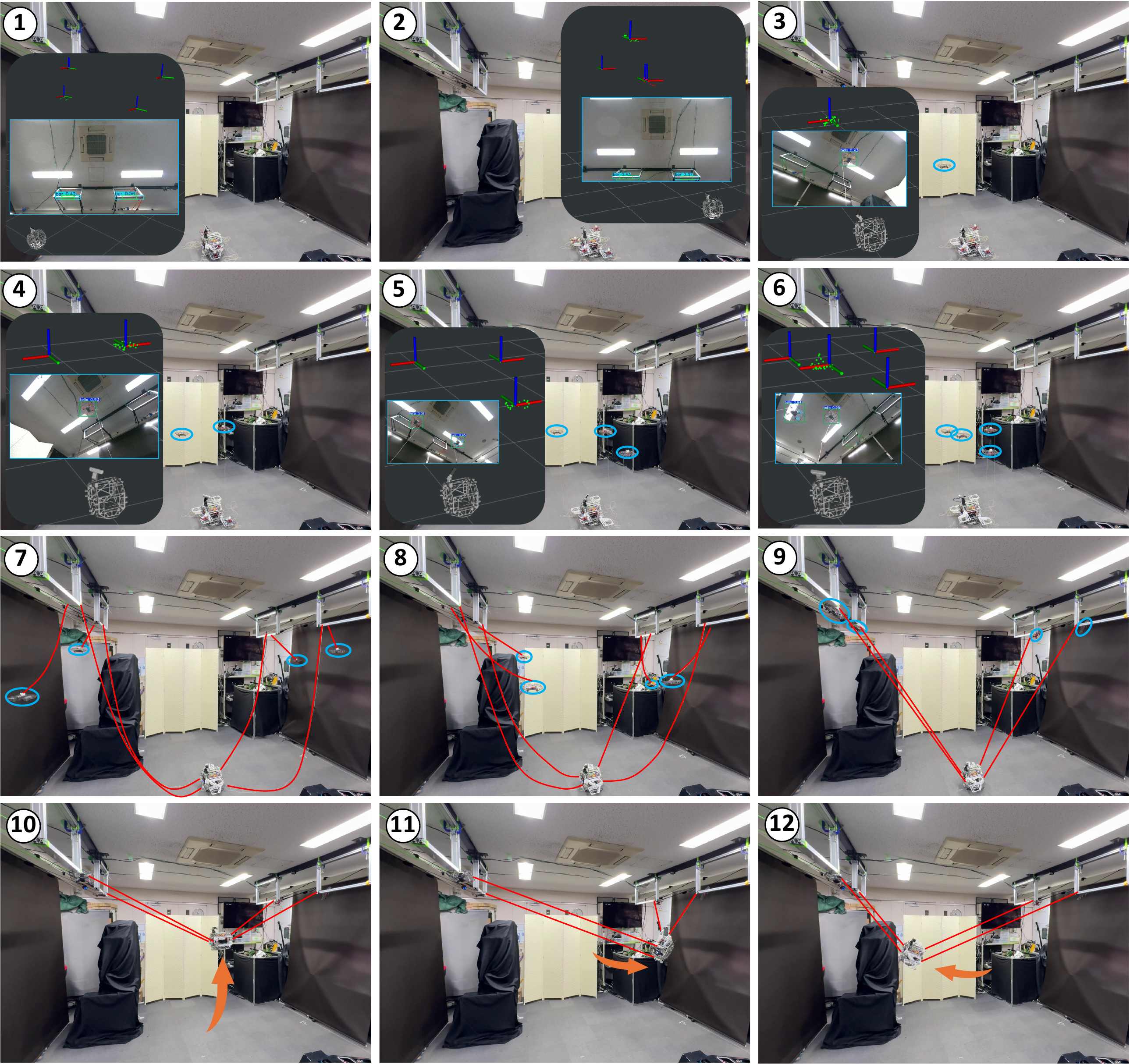}
      \vspace{-5.0ex}
      \caption{
        Scenes from the multiple-wire experiment, showing CubiX and the flying anchors. 
        CubiX simultaneously connects four wires to four frames and drives using those wires.
      }
      \vspace{-3.0ex}
      \label{fig:exp3_pic}
    \end{center}
  \end{figure}
  \begin{figure}[t]
    \begin{center}
      \includegraphics[width=1.0\columnwidth]{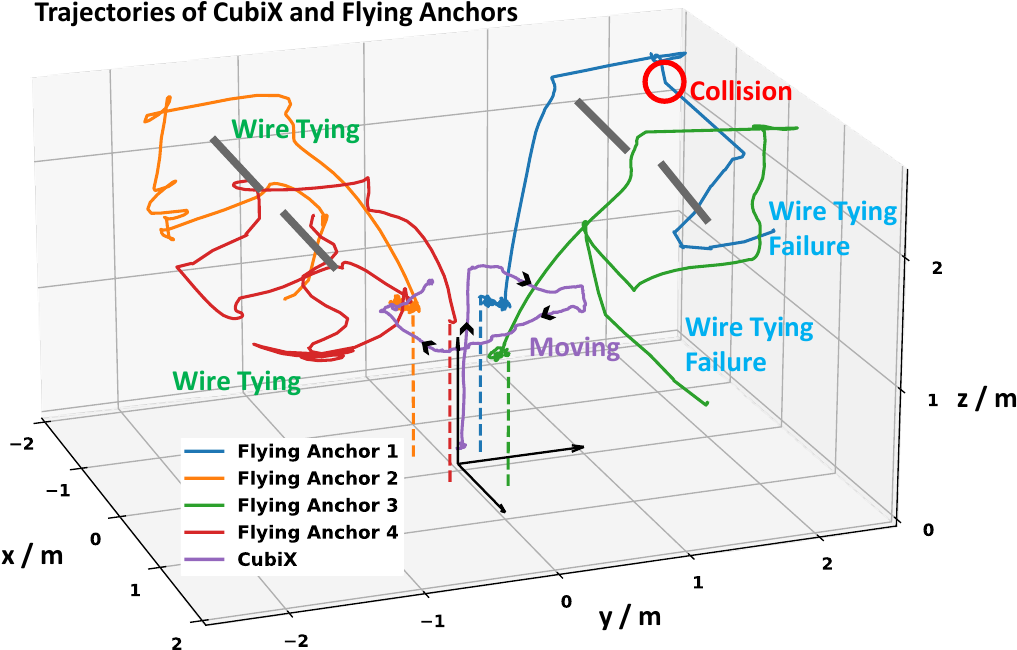}
      \vspace{-5.0ex}
      \caption{
        Trajectories of CubiX and the flying anchors in the multiple-wire experiment. 
        The flying anchors fly to the recognized frames, 
        after which CubiX drives up, down, forward, and backward by adjusting the tension in the wires.
      }
      \vspace{-5.0ex}
      \label{fig:exp3_data}
    \end{center}
  \end{figure}

  To evaluate the ability to handle multiple wires simultaneously, 
  we conducted an experiment in which four wires were attached to a frame in the environment, 
  and CubiX was driven by winding those wires. 
  \figref{fig:exp3_pic} shows CubiX attaching four wires to the environment and driving with them, 
  and \figref{fig:exp3_data} shows the trajectories of CubiX and the flying anchors. 

  In \figref{fig:exp3_pic}\ctext{1}, 
  CubiX recognizes two frames in the environment using its camera and computes their positions and orientations. 
  In \figref{fig:exp3_pic}\ctext{2}, 
  CubiX recognizes the other two frames, which are located on the opposite side, and computes their positions and orientations. 
  In \figref{fig:exp3_pic}\ctext{3}--\ctext{6}, each flying anchor is launched one by one, 
  and CubiX recognizes each anchor's position via its camera. 
  Then, in \figref{fig:exp3_pic}\ctext{7}--\ctext{8}, the flying anchors attach wires to the frames 
  by following the generated paths around the recognized frames. 
  However, as noted in \figref{fig:exp3_data}, the two flying anchors on the right side fail 
  to perform the proposed wire-attachment procedure correctly, 
  ending up physically hooking their anchors directly onto the frame. 
  In \figref{fig:exp3_pic}\ctext{9}, CubiX applies tension to all four wires, 
  thereby securing them via the flying anchors.
  In \figref{fig:exp3_pic}\ctext{10}--\ctext{12}, CubiX increases the tension on the wires, drives, and moves into the air. 
  Additionally, by adjusting the tension of the multiple wires, 
  CubiX was also observed to move forward, backward, left, and right. 
  These results demonstrate that by using the proposed system, 
  multiple wire attachment points can be recognized in the environment, 
  multiple flying anchors can be controlled simultaneously, and multiple wires can be attached to the environment.

  There are two possible causes for a flying anchor's failure to attach its wire.
  First is the recognition error of the anchors by CubiX's camera. 
  Based on the trajectories shown in \figref{fig:exp3_data}, 
  one flying anchor that failed to attach the wire appears not to contact the frame, 
  but in reality it does so where the red circle in \figref{fig:exp3_data} indicates contact. 
  In our experiments, the flying anchors were observed by CubiX's camera only immediately after takeoff; 
  re-observing the anchor at a moment when it is at rest 
  could reduce this error. 
  Second is that the trajectory planning does not account for the state of the wires. 
  While the path was correctly followed by one of the failing anchors, 
  the operation needed to cross wires during attachment was not executed correctly. 
  The feedforward trajectory planning assumes that the wire is sufficiently long and hangs vertically downward 
  from the start point, but in this experiment, the wire had been unwound to an insufficient length, 
  so it was located elsewhere and thus prevented the wire-crossing operation. 
  We believe these issues can be mitigated by explicitly considering the state of the wires, 
  taking into account both the current wire length and the flying anchor's position.
}%
{%
  \begin{figure}[t]
    \begin{center}
      \includegraphics[width=1.0\columnwidth]{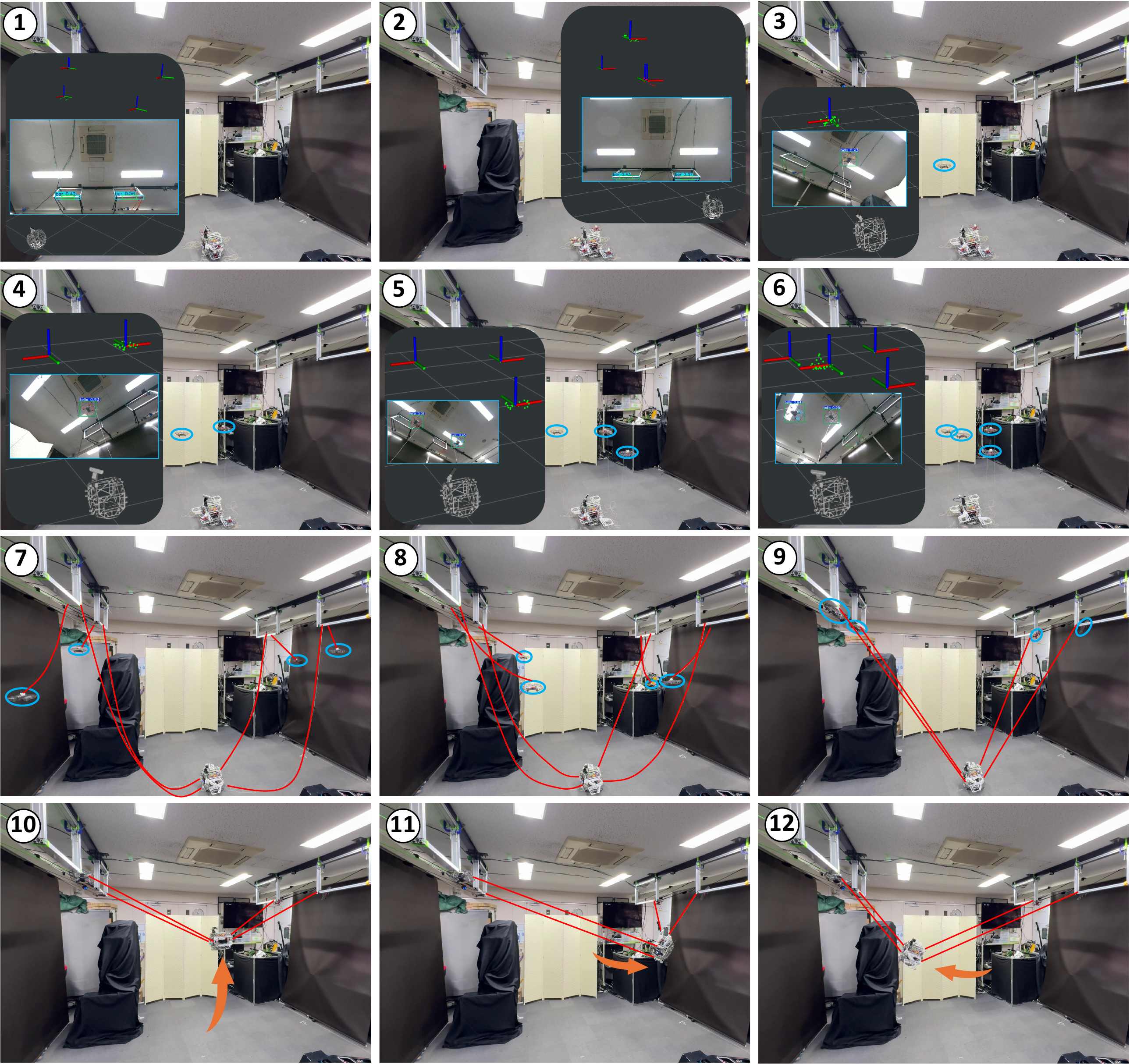}
      \vspace{-5.0ex}
      \caption{複数ワイヤ実験におけるCubiXと飛行アンカーの様子．
      CubiXが4本のワイヤを同時に4つのフレーム接続し，それらのワイヤを用いて駆動する．
      }
      \vspace{-4.0ex}
      \label{fig:exp3_pic}
    \end{center}
  \end{figure}
  \begin{figure}[t]
    \begin{center}
      \includegraphics[width=1.0\columnwidth]{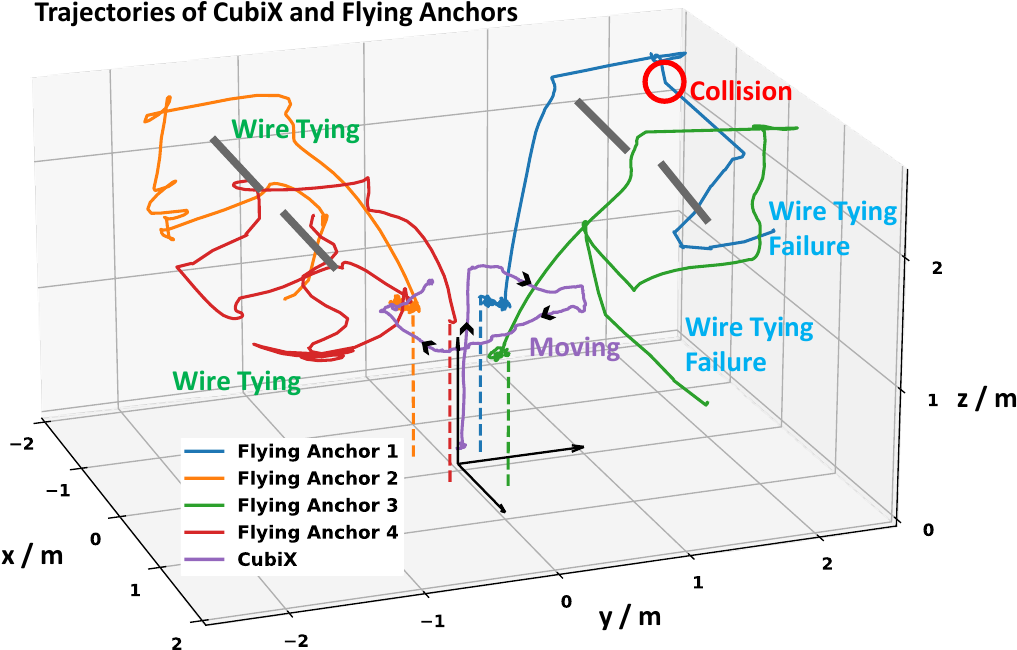}
      \vspace{-5.0ex}
      \caption{複数ワイヤ実験におけるCubiXと飛行アンカーの軌跡．
      認識した4つのフレームに向かって飛行アンカーを制御し，接続したワイヤでCubiXが上下前後左右に駆動したことがわかる．
      }
      \vspace{-5.0ex}
      \label{fig:exp3_data}
    \end{center}
  \end{figure}

  複数本のワイヤを同時に扱うことができるかを評価するため，
  4本のワイヤを環境であるフレームに接続し，それを巻き取って駆動する実験を行った．
  CubiXが環境に4本のワイヤを接続し，そのワイヤで駆動する様子を\figref{fig:exp3_pic}に，
  その際のCubiXと飛行アンカーの軌跡を\figref{fig:exp3_data}に示す．
  \figref{fig:exp3_pic}\ctext{1}で，環境に存在する２つのフレームをCubiXのカメラで認識し，その位置と方向を計算した．
  \figref{fig:exp3_pic}\ctext{2}で，上記フレームとは反対側にあるもう２つのフレームをCubiXのカメラで認識し，
  その位置と方向を計算した．
  \figref{fig:exp3_pic}\ctext{3}--\ctext{6}で，1台ずつ飛行アンカーを離陸させた後，
  各飛行アンカーをCubiXのカメラから認識し， 各飛行アンカーの世界座標系の位置を取得した．
  \figref{fig:exp3_pic}\ctext{7}--\ctext{8}で，認識したフレームの周りに生成した軌跡を各飛行アンカーが追従することで，
  ワイヤをフレームに結びつけた．しかし，\figref{fig:exp3_data}に記される通り，右側の2台の飛行アンカーについては，
  本研究で提案した結びつけに満たない形でアンカーがフレームに直接引っ掛かる形になってる．
  \figref{fig:exp3_pic}\ctext{9}で，CubiXが全てのワイヤに張力をかけることで，フレームに結ばれたワイヤを飛行アンカーで固定した．
  \figref{fig:exp3_pic}\ctext{10}--\ctext{12}で，ワイヤにかけた張力を増やすことで駆動し，CubiXが空中へ移動した．
  また，複数本のワイヤの張力を調節することで前後左右に移動することも確認した．
  これにより，本研究のシステムを用いて，環境から複数ヶ所のワイヤ接続箇所を認識し，
  複数台の飛行アンカーを同時に制御して，複数本のワイヤを環境に接続できることが示された．

  右側の飛行アンカー2台はワイヤの結びつけに失敗は2つの原因が考えられる．
  1つ目に，CubiXカメラによる飛行アンカーの認識誤差である．
  ワイヤの結びつけに失敗した2台の飛行アンカーの内，奥のものに関して，
  \figref{fig:exp3_data}の軌跡を見ると，飛行アンカーはフレームに接触していないが，実際には
  \figref{fig:exp3_data}の赤丸で示した部分で接触している．
  本研究の実験では，飛行アンカーが離陸した直後のみでCubiXのカメラによる飛行アンカーの観測を行ったが，
  ワイヤ結びつけ開始前などの飛行アンカーが停止したタイミングで観測を再び行うことで，観測誤差を小さくできると考えられる．
  ２つ目に，軌跡計画におけてワイヤの状態を考慮していないことである．
  ワイヤの結びつけに失敗した2台の飛行アンカーの内，手前のものに関して，
  正しく経路に追従したものの，ワイヤを結ぶ過程で必要なワイヤ交差させる動作に不備がある．
  フィードフォワード的に軌跡を計算する際には，ワイヤは十分長く，経路始点の鉛直下向きに垂れ下がっていることが仮定されているが，
  本実験では，予め巻き出したワイヤの長さ不足により仮定とは別の場所にワイヤがあり，ワイヤ同士を交差させる動作が実現されなかった．
  これは，現在のワイヤ長さと飛行アンカーの位置からワイヤ状態を陽に考慮することで改善されると考えられる．
}%

\section{Conclusion} \label{sec:conclusion}
\switchlanguage%
{%
  We developed a system enabling the wire-driven robot CubiX 
  to autonomously attach wires—originating from itself—to the environment. 
  We conducted three experiments: 
  attaching a single wire to a bar located at the top of a cliff and climbing by winding that wire; 
  attaching a single wire to a tree branch in an outdoor environment and driving via wire winding; 
  and attaching four wires simultaneously to the environment and driving with them. 
  CubiX uses its onboard RGB-D camera to recognize potential wire-attachment points in the environment, 
  plans a path for tying a wire to those points, 
  and controls the flying anchors (which combine drones and anchor mechanisms) to follow that path,
  thereby autonomously attaching the wire to the environment. 
  By also recognizing the flying anchors using the RGB-D camera on CubiX, 
  we constructed a system in which multiple small flying anchors with low computational resources 
  can be controlled simultaneously. 
  Through these experiments, 
  we demonstrated that CubiX can autonomously recognize the environment and attach its wires in both indoor and outdoor settings. 
  We also demonstrated that CubiX can connect multiple wires to the environment simultaneously.

  As future work, we envision a more practical framework in which, given a specific task and environment, 
  the system recognizes potential wire-attachment points, explores wire configurations that would complete the task, 
  and then uses the current system to set up those wires. 
  Such development would allow wire-driven robots to perform concrete tasks in a variety of real-world environments.
}%
{%
  本研究では，
  ワイヤ駆動ロボットであるCubiXが自身から伸びるワイヤを自律的に環境に接続するシステムを開発し，
  崖上の環境にワイヤを接続し，それを巻き取ることで駆動して崖を登る実験，
  屋外環境の木にワイヤを接続し駆動する実験，
  4本のワイヤを同時に環境に接続し，それらにより駆動する実験を行った．
  CubiXに搭載したRGBDカメラで環境からワイヤ接続箇所を認識し，
  その箇所にワイヤを結びつける軌跡を計画し，
  ドローンとアンカーを組み合わせた飛行アンカーをその軌跡に従って制御することで，
  自律的にワイヤを環境に接続した．
  飛行アンカーも，CubiXに搭載したRGBDカメラから認識することで，小型・低計算リソースで複数台の
  飛行アンカーが制御できるシステムを構築した．
  これにより，各実験を通して，
  屋内外の環境に問わずCubiXは環境を認識して自律的にワイヤを接続することを実現し，
  複数本のワイヤ同時接続も実現できることが示された．

  今後の展望として，
  具体的なタスクと環境が与えられた際に，環境からワイヤ接続箇所を認識し，
  それらからタスクを完遂できるワイヤ配置を探索し，それを今回のシステムで形成できるといったような，
  より実践的な枠組みが開発できれば，
  ワイヤ駆動ロボットが様々な環境において具体的なタスクをこなせるようになることが考えられる．
}%

{
  \bibliographystyle{IEEEtran}
  \bibliography{bib}

\begin{thebibliography}{10}
\providecommand{\url}[1]{#1}
\csname url@rmstyle\endcsname
\providecommand{\newblock}{\relax}
\providecommand{\bibinfo}[2]{#2}
\providecommand\BIBentrySTDinterwordspacing{\spaceskip=0pt\relax}
\providecommand\BIBentryALTinterwordstretchfactor{4}
\providecommand\BIBentryALTinterwordspacing{\spaceskip=\fontdimen2\font plus
\BIBentryALTinterwordstretchfactor\fontdimen3\font minus
  \fontdimen4\font\relax}
\providecommand\BIBforeignlanguage[2]{{%
\expandafter\ifx\csname l@#1\endcsname\relax
\typeout{** WARNING: IEEEtran.bst: No hyphenation pattern has been}%
\typeout{** loaded for the language `#1'. Using the pattern for}%
\typeout{** the default language instead.}%
\else
\language=\csname l@#1\endcsname
\fi
#2}}

\bibitem{5979640}
A.~Capua, A.~Shapiro, and S.~Shoval, ``Spiderbot: A cable suspended mobile
  robot,'' in \emph{2011 IEEE International Conference on Robotics and
  Automation}, 2011, pp. 3437--3438.

\bibitem{10619924}
J.~Quan, M.~Zhu, and D.~Hong, ``Re-examining climbing robots: Design and
  performance of a lightweight, low-cost robot with a highly extendable limb,''
  in \emph{2024 6th International Conference on Reconfigurable Mechanisms and
  Robots (ReMAR)}, 2024, pp. 409--416.

\bibitem{10296064}
T.~Nishio, M.~Zhao, K.~Okada, and M.~Inaba, ``Design, control, and motion
  planning for a root-perching rotor-distributed manipulator,'' \emph{IEEE
  Transactions on Robotics}, vol.~40, pp. 660--676, 2024.

\bibitem{titan_xi_slope}
R.~Hodoshima, T.~Doi, Y.~Fukuda, S.~Hirose, T.~Okamoto, and J.~Mori,
  ``Development of titan xi: a quadruped walking robot to work on slopes,'' in
  \emph{Proceedings of the 2004 IEEE/RSJ International Conference on
  Intelligent Robots and Systems}, vol.~1, 2004, pp. 792--797.

\bibitem{8794265}
T.~Miki, P.~Khrapchenkov, and K.~Hori, ``{UAV/UGV Autonomous Cooperation: UAV
  assists UGV to climb a cliff by attaching a tether},'' in \emph{Proceedings
  of the 2019 IEEE International Conference on Robotics and Automation}, 2019,
  pp. 8041--8047.

\bibitem{9517518}
T.~Kominami, H.~Paul, R.~Miyazaki, B.~Sumetheeprasit, R.~Ladig, and
  K.~Shimonomura, ``{Active Tethered Hook: Heavy Load Movement using Hooks that
  Move Actively with Micro UAVs and Winch System},'' in \emph{Proceedings of
  the 2021 IEEE/ASME International Conference on Advanced Intelligent
  Mechatronics}, 2021, pp. 264--269.

\bibitem{10375200}
S.~Yuzaki, A.~Miki, M.~Bando, S.~Yoshimura, T.~Suzuki, K.~Kawaharazuka,
  K.~Okada, and M.~Inaba, ``Fusion of body and environment with movable
  carabiners for wire-driven robots toward expansion of physical
  capabilities,'' in \emph{Proceedings of the 2023 IEEE-RAS International
  Conference on Humanoid Robots}, 2023, pp. 1--7.

\bibitem{5509299}
J.-P. Merlet and D.~Daney, ``A portable, modular parallel wire crane for rescue
  operations,'' in \emph{Proceedings of the 2010 IEEE International Conference
  on Robotics and Automation}, 2010, pp. 2834--2839.

\bibitem{merlet2010marionet}
J.-P. Merlet, ``Marionet, a family of modular wire-driven parallel robots,'' in
  \emph{Advances in Robot Kinematics: Motion in Man and Machine: Motion in Man
  and Machine}.\hskip 1em plus 0.5em minus 0.4em\relax Springer, 2010, pp.
  53--61.

\bibitem{cubix}
S.~Inoue, K.~Kawaharazuka, T.~Suzuki, S.~Yuzaki, K.~Okada, and M.~Inaba,
  ``{Overcoming Physical Limitations Utilizing the Surrounding Environment with
  a Wire-Driven Multipurpose Robot},'' \emph{Advanced Robotics Research}, p.
  202400021.

\bibitem{TelloEDU}
``{Tello EDU},'' \url{https://www.ryzerobotics.com/tello-edu}, {Accessed:
  2025-3-1}.

\bibitem{7780460}
J.~Redmon, S.~Divvala, R.~Girshick, and A.~Farhadi, ``You only look once:
  Unified, real-time object detection,'' in \emph{Proceedings of the 2016 IEEE
  International Conference on Computer Vision and Pattern Recognition}, 2016,
  pp. 779--788.

\bibitem{boris_sekachev_2020_4009388}
\BIBentryALTinterwordspacing
B.~Sekachev, N.~Manovich, M.~Zhiltsov, \emph{et~al.}, ``opencv/cvat: v1.1.0,''
  Aug. 2020. [Online]. Available: \url{https://doi.org/10.5281/zenodo.4009388}
\BIBentrySTDinterwordspacing

\bibitem{jolliffe2002principal}
I.~T. Jolliffe, \emph{Principal Component Analysis}, 2nd~ed.\hskip 1em plus
  0.5em minus 0.4em\relax Springer, 2002.

\bibitem{5980567}
R.~B. Rusu and S.~Cousins, ``{3D is here: Point Cloud Library (PCL)},'' in
  \emph{2011 IEEE International Conference on Robotics and Automation}, 2011,
  pp. 1--4.

\bibitem{1360855570047666048}
R.~E. Kalman, ``A new approach to linear filtering and prediction problems,''
  \emph{Journal of Basic Engineering}, vol.~82, no.~1, pp. 35--45, 03 1960.

\end{thebibliography}
}

\end{document}